\documentclass[journal]{IEEEtran}
\usepackage{amsmath,amsfonts,textcomp,gensymb,amssymb, amsthm}
\usepackage{algorithmic,algorithm}
\usepackage{tabularx,booktabs}
\usepackage{graphicx,caption, subcaption}
\usepackage{colortbl}
\usepackage[table]{xcolor}
\definecolor{NavyBlue}{RGB}{0, 49, 83}
\usepackage{array}
\usepackage{xcolor}
\usepackage{textcomp}
\usepackage{stfloats}
\usepackage{url}
\usepackage{verbatim}
\usepackage{graphicx}
\usepackage{cite}

\makeatletter
\newenvironment{breakablealgorithm}
{
	\begin{center}
		\refstepcounter{algorithm}
		\hrule height.8pt depth0pt \kern2pt
		\renewcommand{\caption}[2][\relax]{
			{\raggedright\textbf{\ALG@name~\thealgorithm} ##2\par}%
			\ifx\relax##1\relax 
			\addcontentsline{loa}{algorithm}{\protect\numberline{\thealgorithm}##2}%
			\else 
			\addcontentsline{loa}{algorithm}{\protect\numberline{\thealgorithm}##1}%
			\fi
			\kern2pt\hrule\kern2pt
		}
	}{
		\kern2pt\hrule\relax
	\end{center}
}
\makeatother

\begin{document}
\title{Coordinated guidance and control for \\ multiple parafoil system landing}

\author{Zhenyu Wei, Zhijiang Shao, Lorenz T. Biegler
\thanks{Zhenyu Wei and Zhijiang Shao are with College of Control Science and Engineering, Zhejiang University, Hangzhou, China.}
\thanks{Lorenz T. Biegler is with Department of Chemical Engineering, Carnegie Mellon University, Pittsburgh, USA.}}

\maketitle

\begin{abstract}
  Multiple parafoil landing is an enabling technology for massive supply delivery missions. However, it is still an open question to design a collision-free, computation-efficient guidance and control method for unpowered parafoils. To address this issue, this paper proposes a coordinated guidance and control method for multiple parafoil landing. First, the multiple parafoil landing process is formulated as a trajectory optimization problem. Then, the landing point allocation algorithm is designed to assign the landing point to each parafoil. In order to guarantee flight safety, the collision-free trajectory replanning algorithm is designed. On this basis, the nonlinear model predictive control algorithm is adapted to leverage the nonlinear dynamics model for trajectory tracking. Finally, the parafoil kinematic model is utilized to reduce the computational burden of trajectory calculation, and kinematic model is updated by the moving horizon correction algorithm to improve the trajectory accuracy. Simulation results demonstrate the effectiveness and computational efficiency of the proposed coordinated guidance and control method for the multiple parafoil landing.
\end{abstract}


\section{Introduction}
\IEEEPARstart{T}{he} parafoil system, also known as the precision aerial delivery system, is an unmanned aerial vehicle (UAV) that can accurately and consistently deliver cargo to a designated location. Due to the advantage of load capability and flight controllability, the parafoil system has been widely applied in supply delivery \cite{[1]carter2007autonomous} and flight vehicle recovery \cite{[2]strahan2003testing}. The development from single parafoil to multiple parafoil drop has received great attention in recent years. The airdrop of multiple parafoils enables the rapid delivery of a massive amount of supplies to the designated place, which would benefit, for instance, humanitarian supply delivery missions \cite{[3]sun2021distributed}.

The challenge for multiple parafoil landing has two parts. One is that the parafoil system, due to its low speed and unpowered property, is susceptible to wind disturbances and has limited control over its flight trajectory \cite{[4]figueroa2021landing}. Another is that the parafoil swarm should avoid collision during flight and land precisely at the desired landing area \cite{[5]chen2020consensus}. As a result, the multiple parafoil landing requires an efficient guidance and control approach for coordination.

Current approaches for parafoil system landing generally fall into three categories. Waypoint-based algorithms generate a sequence of waypoints from the exit position of the energy management phase to the target landing position and design the controller to track these waypoints \cite{[6]yiming2023adaptive}. Maneuver-based algorithms generate a reference glide slope to the target and perform a sequence of maneuvers to maintain the glide slope \cite{[7]calise2008swarming}. Path-based algorithms generate a continuous reference trajectory to the landing point and utilize the controller to track the generated trajectory \cite{[8]sun2022trajectory,[9]zheng2023sideslip}. The reference trajectory can be smooth curves, such as Bezier curve \cite{[10]fowler2014bezier} and Dubins path \cite{[11]rademacher2009flight}, or the optimal trajectory obtained by trajectory optimization algorithms \cite{[12]zhang2013multi,[13]leeman2023autonomous}. The control method for tracking the reference trajectory can be linear model predictive control (LMPC) \cite{[4]figueroa2021landing}, active disturbance rejection control \cite{[14]luo2019accurate}, sliding mode control \cite{[8]sun2022trajectory}, and proportional-integral-derivative control \cite{[15]tao2017dynamic}. In previous literature, the steady-state gliding condition \cite{[16]ward2013adaptive} is assumed to reduce the high-order parafoil dynamic model to the three-degree-of-freedom (3-DOF) kinematic model, which was utilized to facilitate the design of parafoil trajectory planning and tracking algorithms. However, this gliding condition is seldom satisfied during the flight of the parafoil system. Hence, there exists a model deviation between the actual parafoil system and the guidance and control algorithm. In real flight, this model deviation may cause significant trajectory errors and eventually contribute to the terminal landing error \cite{[15]tao2017dynamic}.

Previous literature on multiple parafoil system landing can be divided into three types based on the strategy to avoid collision. The first type uses biology behavior-based mechanisms \cite{[7]calise2008swarming} or designed flocking rules \cite{[17]rosich2012coupling} to coordinate the parafoils. It relies on detecting the relative distance to trigger the collision avoidance maneuver, which is vulnerable because this short-sighted planning may not recognize a collision until it is too late. The second type leverages the consensus-based formation method, such as virtual structure \cite{[18]chen2019virtual} and leader-follower theory \cite{[3]sun2021distributed,[5]chen2020consensus}, to keep the formation of multiple parafoil until touch down. However, it is assumed in these papers that the parafoils are equipped with a propulsion system to maintain steady and tight flight formation, which does not apply to normal airdrop missions. The third type constructs the multi-parafoil landing as a centralized trajectory optimization problem \cite{[19]kaminer2005coordinated}, which is solved by the genetic algorithm \cite{[20]qi2019multi} or the pseudospectral method \cite{[21]luo2016trajectory} to obtain the guidance trajectories. Although collision avoidance is explicitly considered as the path constraint in the trajectory optimization problem, this centralized framework suffers from a computational burden and is not suitable for online operation. Besides, the landing points are pre-decided for each parafoil, which lacks an effective allocation method to improve overall performance and adaptability.

This paper aims to design a coordinated guidance and control method for multiple parafoil system landing, which would benefit the massive supply delivery mission. To this end, we outline the parafoil system dynamics and formulate the multiple parafoil landing trajectory optimization problem. The landing point allocation algorithm is designed to assign the landing point to each parafoil. Then, the collision-free trajectory replanning algorithm is designed to avoid trajectory collision. The nonlinear model predictive control (NMPC) method is introduced to leverage the nonlinear dynamic model for trajectory tracking. To reduce the computational burden, the parafoil kinematic model is utilized, and the moving horizon correction algorithm is designed to update the kinematic model. The main contributions of this paper are:
\begin{enumerate}
	\item The landing points are rapidly evaluated by each parafoil via the NLP sensitivity analysis. On this basis, an assignment problem is formulated and solved by the Hungarian algorithm to optimally allocate the landing point to each parafoil.
	\item The collision avoidance sub-problems are solved in coordination to generate collision-free trajectories in a decoupled manner.
	\item The NMPC algorithm is adapted to track the guidance trajectory based on parafoil dynamic model, which reduces the influence of disturbances.
	\item The parafoil kinematic model is updated by the moving horizon estimation algorithm to increase the guidance trajectory accuracy.
	\item Simulation results demonstrate the effectiveness of the proposed coordinated guidance and control approach for multiple parafoil landing.
\end{enumerate}

The rest of the paper is organized as follows. Section \ref{Sec2} constructs the multiple parafoil landing problem. Section \ref{Sec3} introduces the coordinated guidance and control method. Section \ref{Sec4} presents the simulation results and discussion. Section \ref{Sec5} concludes the paper.

\section{Multiple parafoil system landing problem formulation}
\label{Sec2}
In this section, the trajectory optimization problem for the landing of multiple parafoil is formulated. First, the parafoil system dynamics are outlined. Then, the overall trajectory optimization problem is established.
\subsection{Parafoil system dynamics}
\begin{figure}[!ht]
\centering
\includegraphics[width=1.8in]{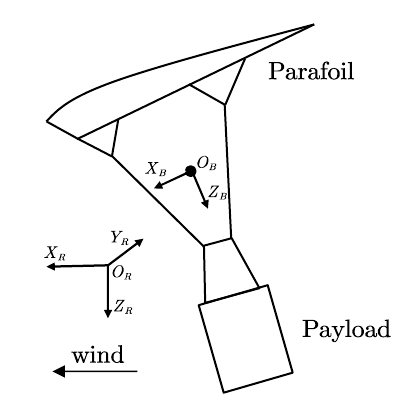}\\
\caption{Reference frames of the parafoil system.}
\label{Fig1}
\end{figure}

In order to facilitate the discussion in this paper, the ground reference frame and the parafoil-body-fixed reference frame are defined. The ground frame (referred to as $O_RX_RY_RZ_R$) sets the center of landing area as its origin, with its $O_RX_R$ axis pointing along the direction of the known constant wind. The parafoil body frame (referred to as $O_BX_BY_BZ_B$) uses the mass center of the parafoil system as its origin. The reference frames are shown in Fig. \ref{Fig1}.

The translational and rotational movements of the parafoil system are expressed as follows:

\begin{equation}
\begin{gathered}
\label{equation: 1}
\begin{array}{l}
	\dot{\boldsymbol{r}}_R=\left( \boldsymbol{C}_{R}^{B} \right) ^{\mathrm{T}}\boldsymbol{v}_B\\
	\dot{\boldsymbol{o}}_{B\gets R}=\boldsymbol{C}_{\omega}\boldsymbol{\omega }_B\\
\end{array}
\end{gathered}
\end{equation}
\noindent where $\boldsymbol{r}_R=\left[ \begin{matrix}
	x&		y&		z\\
\end{matrix} \right] ^{\mathrm{T}}$ denotes the parafoil position vector in the ground frame; $\boldsymbol{o}_{B\gets R}=\left[ \begin{matrix}
	\phi&		\theta&		\psi\\
\end{matrix} \right] ^{\mathrm{T}}$ denotes the relative attitude of the parafoil system from the ground frame to the body frame; $\phi$, $\theta$, and $\psi$ refer to the roll, pitch, and yaw angle, respectively; $\boldsymbol{v}_B\in \mathbb{R} ^3$ and $\boldsymbol{\omega }_B\in \mathbb{R} ^3$ refer to the velocity and angular velocity in the parafoil body frame; $\boldsymbol{C}_{R}^{B}\in \mathbb{R} ^{3\times 3}$ represents the transformation matrix from the ground frame to the parafoil body frame, and its transpose matrix represents the opposite; $\boldsymbol{C}_{\omega}\in \mathbb{R} ^{3\times 3}$ represents the transformation matrix for the angular velocity.

The parafoil system considered in this paper adopts a four-point rigging harness connection, which is common for heavy systems. When utilizing the four-point connection, the roll and pitch relative motions are prohibited, and the yaw relative movement has limited influence on the trajectory of the parafoil system \cite{[22]yakimenko2015precision}. We assume that the parafoil and the payload are connected as a rigid body, thus the 6-DOF parafoil system model is considered as follows:

\begin{equation}
\begin{gathered}
\label{equation: 2}
\boldsymbol{A}\left[ \begin{array}{c}
	\dot{\boldsymbol{v}}_B\\
	\dot{\boldsymbol{\omega}}_B\\
\end{array} \right] =\left[ \begin{array}{c}
	\boldsymbol{F}_a+\boldsymbol{F}_g\\
	\boldsymbol{M}_a\\
\end{array} \right] +\boldsymbol{B}\left[ \begin{array}{c}
	\boldsymbol{v}_B\\
	\boldsymbol{\omega }_B\\
\end{array} \right] +\boldsymbol{C}
\\
\begin{array}{l}
	\begin{array}{l}
	\boldsymbol{A}_{11}=m\boldsymbol{I}_{3\times 3}+\boldsymbol{m}_{a}^{\prime},\boldsymbol{A}_{12}=-\boldsymbol{m}_{a}^{\prime}\left[ \boldsymbol{r}_{B}^{a}\times \right]\\
	\boldsymbol{A}_{21}=\left[ \boldsymbol{r}_{B}^{a}\times \right] \boldsymbol{m}_{a}^{\prime},\boldsymbol{A}_{22}=\boldsymbol{J}+\boldsymbol{J}_{a}^{\prime}-\left[ \boldsymbol{r}_{B}^{a}\times \right] \boldsymbol{m}_{a}^{\prime}\left[ \boldsymbol{r}_{B}^{a}\times \right]\\
\end{array}\\
	\begin{array}{l}
	\boldsymbol{B}_{11}=-\left[ \boldsymbol{\omega }_B\times \right] \left( m\boldsymbol{I}_{3\times 3}+\boldsymbol{m}_{a}^{\prime} \right)\\
	\boldsymbol{B}_{12}=\left[ \boldsymbol{\omega }_B\times \right] \boldsymbol{m}_{a}^{\prime}\left[ \boldsymbol{r}_{B}^{a}\times \right]\\
	\boldsymbol{B}_{21}=-\left[ \boldsymbol{r}_{B}^{a}\times \right] \left[ \boldsymbol{\omega }_B\times \right] \boldsymbol{m}_{a}^{\prime}\\
	\boldsymbol{B}_{22}=-\left( \left[ \boldsymbol{\omega }_B\times \right] \left( \boldsymbol{J}+\boldsymbol{J}_{a}^{\prime} \right) -\left[ \boldsymbol{r}_{B}^{a}\times \right] \left[ \boldsymbol{\omega }_B\times \right] \boldsymbol{m}_{a}^{\prime}\left[ \boldsymbol{r}_{B}^{a}\times \right] \right)\\
\end{array}\\
\end{array}
\end{gathered}
\end{equation}
\noindent where $m$ and $\boldsymbol{J}$ denote the total mass and total rotational inertia of the parafoil system; $\boldsymbol{m}_{a}^{\prime}$ and $\boldsymbol{J}_{a}^{\prime}$ refer to the apparent mass and the apparent inertia term; $\boldsymbol{F}_a$ and $\boldsymbol{F}_g$ denote the sum of aerodynamic force and the sum of gravity force on the system; $\boldsymbol{M}_a$ denotes the aerodynamic moment of the system; $\boldsymbol{r}_{B}^{a}$ represent the position vector from the center of gravity to that of the apparent mass. The skew-symmetric matrix $\left[ \boldsymbol{\xi }\times \right]$ is the cross-product matrix for vector $\boldsymbol{\xi }\in \mathbb{R} ^3$. The details of the dynamics of the parafoil system are referred to \cite{[23]wei2024dynamic}.

The complexity of parafoil dynamics mainly lies in two parts. One is that the apparent mass and the apparent inertia cause the coupling of the velocity and the angular velocity. Another is that the aerodynamic terms are depicted by nonlinear functions containing the aerodynamic angle, velocity, angular velocity, and control command.

\subsection{Formulated trajectory optimization problem}
Combining (\ref{equation: 1})-(\ref{equation: 2}), the 6-DOF dynamic model for \textit{i}th parafoil system is established as $\dot{\boldsymbol{x}}^i\left( t \right) =\boldsymbol{f}^i\left( \boldsymbol{x}^i\left( t \right) ,\boldsymbol{u}^i\left( t \right) \right)$, where the state and control variables are expressed as $\boldsymbol{x}^i\left( t \right) =\left[ \begin{matrix}
	\boldsymbol{r}_{R}^{i}&		\boldsymbol{o}_{B\gets R}^{i}&		\boldsymbol{v}_{B}^{i}&		\boldsymbol{\omega }_{B}^{i}\\
\end{matrix} \right] ^{\mathrm{T}}$ and $\boldsymbol{u}^i\left( t \right) =\left[ \begin{matrix}
	\delta _{a}^{i}&		\delta _{s}^{i}\\
\end{matrix} \right] ^{\mathrm{T}}$.

The actual control quantities of the parafoil system are left flap deflection $\delta _{l}^{i}$ and right flap deflection $\delta _{r}^{i}$, which are constrained within the bound $\left[ 0, \delta _{\max} \right]$. In order to facilitate the formulation of the trajectory optimization problem, the equivalent control quantities $\delta _{a}^{i}=\delta _{r}^{i}-\delta _{l}^{i}$ and $\delta _{s}^{i}=\min \left\{ \delta _{l}^{i},\delta _{r}^{i} \right\}$ are utilized. Therefore, the control constraints are transformed into the equivalent form as follows:

\begin{equation}
\begin{gathered}
\label{equation: 3}
\begin{array}{l}
	0\leqslant \delta _{s}^{i}\left( t \right)\\
	\delta _{s}^{i}\left( t \right) +\delta _{a}^{i}\left( t \right) \leqslant \delta _{\max}\\
	\delta _{s}^{i}\left( t \right) -\delta _{a}^{i}\left( t \right) \leqslant \delta _{\max}\\
\end{array}
\end{gathered}
\end{equation}

The \textit{i}th parafoil should keep a safe distance with other parafoils to avoid collision, which is defined by the following constraint:

\begin{equation}
\begin{gathered}
\label{equation: 4}
\left\| \boldsymbol{r}_{R}^{i}\left( t \right) -\boldsymbol{r}_{R}^{j}\left( t \right) \right\| ^2\geqslant \left( d_s \right) ^2,j=1,2,..,N,j\ne i
\end{gathered}
\end{equation}
\noindent where $d_s$ is the safety distance required between parafoils to avoid collision risk.

The initial state is set as $\boldsymbol{x}^i\left( 0 \right) =\boldsymbol{x}_{init}^{i}$ corresponding to the status of the \textit{i}th parafoil at the start time. The terminal position constraint is defined as:

\begin{equation}
\begin{gathered}
\label{equation: 5}
\begin{array}{l}
  \left( x_{f}^{i}\left( t_{f}^{i} \right) -x_f \right) ^2+\left( y_{f}^{i}\left( t_{f}^{i} \right) -y_f \right) ^2\leqslant \left( d_f \right) ^2,z^i\left( t_{f}^{i} \right) =z_f
\end{array}
\end{gathered}
\end{equation}
\noindent where $\boldsymbol{r}_{R}^{i}=\left[ \begin{matrix}
	x^i&		y^i&		z^i\\
\end{matrix} \right] ^{\mathrm{T}}$; $z_f$ denotes the terminal landing altitude; $t_{f}^{i}$ is the terminal time of \textit{i}th parafoil. To facilitate payload recovery, the parafoil system is confined in a target circular area, with the center point $\left( x_f,y_f \right)$ and radius $d_f$.

Besides, the parafoil should land upwind to reduce the influence of wind disturbance. Therefore, the terminal angle is constrained as:

\begin{equation}
\begin{gathered}
\label{equation: 6}
\psi ^i\left( t_{f}^{i} \right) =\alpha _{anti\_wind}
\end{gathered}
\end{equation}

The overall control effort of the landing flight should be minimized, which improves the parafoil systems’ ability to resist wind disturbances and avoid collision. Hence, the objective function is given as follows:

\begin{equation}
\begin{gathered}
\label{equation: 7}
J=\sum\nolimits_{i=1}^N{\int_0^{t_{f}^{i}}{\left\| \boldsymbol{u}^i\left( t \right) \right\| ^2}dt}
\end{gathered}
\end{equation}

Finally, the multiple parafoil landing trajectory optimization problem is formulated as follows:

\begin{equation}
\begin{gathered}
\label{equation: 8}
\begin{array}{l}
	\underset{\boldsymbol{u}^i,t_{f}^{i}}{\min} J=\sum\nolimits_{i=1}^N{\int_0^{t_{f}^{i}}{\left\| \boldsymbol{u}^i\left( t \right) \right\| ^2dt}}\\
	s.t.\,\dot{\boldsymbol{x}}^i\left( t \right) =f^i\left( \boldsymbol{x}^i\left( t \right) ,\boldsymbol{u}^i\left( t \right) \right)\\
	\quad\,\,\,\, 0\leqslant \delta _{s}^{i}\left( t \right) , \delta _{s}^{i}\left( t \right) +\delta _{a}^{i}\left( t \right) \leqslant \delta _{\max},\delta _{s}^{i}\left( t \right) -\delta _{a}^{i}\left( t \right) \leqslant \delta _{\max}\\
	\quad\,\,\,\, \left\| \boldsymbol{r}_{R}^{i}\left( t \right) -\boldsymbol{r}_{R}^{j}\left( t \right) \right\| ^2\geqslant \left( d_s \right) ^2,i=1,...,N-1,j>i\\
	\quad\,\,\,\, \boldsymbol{x}^i\left( 0 \right) =\boldsymbol{x}_{init}^{i}\\
	\quad\,\,\,\, \left( x^i\left( t_{f}^{i} \right) -x_f \right) ^2+\left( y^i\left( t_{f}^{i} \right) -y_f \right) ^2\leqslant \left( d_f \right) ^2,\\
	\quad\,\,\,\, z^i\left( t_{f}^{i} \right) =z_f,\psi ^i\left( t_{f}^{i} \right) =\alpha _{anti\_wind}\\
  \quad\,\,\,\, i=1,...,N\\
\end{array}
\end{gathered}
\end{equation}

The formulated problem (\ref{equation: 8}) can be directly solved by the trajectory optimization method, such as the heuristic method or the pseudospectral method. However, problem (\ref{equation: 8}) is non-convex and contains complex parafoil dynamic constraints and coupled collision avoidance path constraints, resulting in intense computation. Besides, each parafoil system has a different terminal time, and the problem scale increases with the number of parafoils, further increasing the computational difficulty. Moreover, parafoils usually have limited computation and communication capabilities, which makes it impractical to perform such a complex centralized computation.

\section{Coordinated guidance and control method}
\label{Sec3}
This section introduces the coordinated guidance and control method for the multiple parafoil landing. First, the landing point allocation algorithm is designed to assign the landing point to each parafoil. Then, the collision avoidance trajectory is replanned to ensure flight safety. Thereafter, the NMPC controller and moving horizon correction are adapted to track the reference trajectory and update the kinematic model, respectively. Finally, the overall structure of the coordinated guidance and control framework is summarized.
\subsection{Landing point allocation algorithm}
Assuming we have $N$ pre-determined landing points in the landing area, the terminal position constraint in problem (\ref{equation: 8}) can be converted to the fixed-point landing constraint as in the previous parafoil landing literature, which helps to reduce the problem complexity. Hence, the subsequent challenge is how to reasonably allocate the landing point.

To start with, we formulate following fixed-point landing problem to consider the \textit{i}th parafoil to land at the \textit{k}th landing point $\boldsymbol{\theta }^k=\left[ \begin{matrix}
	\theta _{x}^{k}&		\theta _{y}^{k}\\
\end{matrix} \right] ^{\mathrm{T}}$:

\begin{equation}
\begin{gathered}
\label{equation: 9}
\begin{array}{l}
	\underset{\boldsymbol{u}^i,t_{f}^{i}}{\min} \int_0^{t_{f}^{i}}{\left\| \boldsymbol{u}^i\left( t \right) \right\| ^2dt}\\
	s.t. \,\dot{\boldsymbol{x}}^i\left( t \right) =f^i\left( \boldsymbol{x}^i\left( t \right) ,\boldsymbol{u}^i\left( t \right) \right)\\
	\quad\,\,\,\, 0\leqslant \delta _{s}^{i}\left( t \right) , \delta _{s}^{i}\left( t \right) +\delta _{a}^{i}\left( t \right) \leqslant \delta _{\max},\delta _{s}^{i}\left( t \right) -\delta _{a}^{i}\left( t \right) \leqslant \delta _{\max}\\
	\quad\,\,\,\, \boldsymbol{x}^i\left( 0 \right) =\boldsymbol{x}_{init}^{i}\\
	\quad\,\,\,\,  x^i\left( t_{f}^{i} \right) =\theta _{x}^{k},y^i\left( t_{f}^{i} \right) =\theta _{y}^{k}\\
	\quad\,\,\,\, z^i\left( t_{f}^{i} \right) =z_f,\psi ^i\left( t_{f}^{i} \right) =\alpha _{anti\_wind}\\
\end{array}
\end{gathered}
\end{equation}

To ensure the solution feasibility when the terminal conditions are infeasible, we re-write the terminal constraints as the $\ell _1$ penalty terms and augment to the objective function. Hence, problem (\ref{equation: 9}) is reformulated as:

\begin{equation}
\begin{gathered}
\label{equation: 11}
\begin{array}{l}
	\underset{\boldsymbol{u}^i,t_{f}^{i}}{\min}\int_0^{t_{f}^{i}}{\left\| \boldsymbol{u}^i\left( t \right) \right\| ^2dt}+\beta \sum{\left( \boldsymbol{p}+\boldsymbol{n} \right) ^{\mathrm{T}}}\boldsymbol{e}\\
	s.t.\,\dot{\boldsymbol{x}}^i\left( t \right) =f^i\left( \boldsymbol{x}^i\left( t \right) ,\boldsymbol{u}^i\left( t \right) \right)\\
	\quad \,\,\,\,0\leqslant \delta _{s}^{i}\left( t \right) ,\delta _{s}^{i}\left( t \right) +\delta _{a}^{i}\left( t \right) \leqslant \delta _{\max},\delta _{s}^{i}\left( t \right) -\delta _{a}^{i}\left( t \right) \leqslant \delta _{\max}\\
	\quad \,\,\,\,\boldsymbol{x}^i\left( 0 \right) =\boldsymbol{x}_{init}^{i}\\
	\quad \,\,\,\,x^i\left( t_{f}^{i} \right) -\theta _{x}^{k}=p_x-n_x\\
	\quad \,\,\,\,y^i\left( t_{f}^{i} \right) -\theta _{y}^{k}=p_y-n_y\\
	\quad \,\,\,\,z^i\left( t_{f}^{i} \right) -z_f=p_z-n_z\\
	\quad \,\,\,\,\psi ^i\left( t_{f}^{i} \right) -\alpha _{anti\_wind}=p_{\psi}-n_{\psi}\\
	\quad \,\,\,\,\boldsymbol{p}\geqslant 0,\boldsymbol{n}\geqslant 0\\
\end{array}
\end{gathered}
\end{equation}
\noindent where $\boldsymbol{p}=\left[ \begin{matrix}
	p_x&		p_y&		p_z&		p_{\psi}\\
\end{matrix} \right] ^{\mathrm{T}}$, $\boldsymbol{n}=\left[ \begin{matrix}
	n_x&		n_y&		n_z&		n_{\psi}\\
\end{matrix} \right] ^{\mathrm{T}}$, and $e=\left[ 1,...,1 \right] ^{\mathrm{T}}$; $\beta$ is the penalty weight, which is chosen by trial and error to satisfy $\beta \geqslant \left\| \boldsymbol{\lambda }_{\beta} \right\| _{\infty}$; $\boldsymbol{\lambda }_{\beta}$ is the corresponding dual variable for the related constraints \cite{[24]biegler2010nonlinear}.

For \textit{i}th parafoil, problem (\ref{equation: 11}) is solved $N$ times to evaluate the control energy consumption for the \textit{i}th parafoil system to land at \textit{k}th landing point, which is denoted as the $R^i\left( \boldsymbol{\theta }^k \right) =\int_0^{t_{f}^{i}}{\left\| \boldsymbol{u}^i\left( t \right) \right\| ^2dt}$.

To reduce the computation time of repeatedly solving problem (\ref{equation: 11}) with different $\boldsymbol{\theta }^k$, NLP sensitivity analysis is utilized. First, problem (\ref{equation: 11}) with the target area center $\boldsymbol{\theta }^0=\left[ \begin{matrix}
	x_f&		y_f\\
\end{matrix} \right] ^{\mathrm{T}}$ as the landing point is discretized by the simultaneous collocation method \cite{[25]kameswaran2006simultaneous}. The discretized form of problem (\ref{equation: 11}) can be written as the following general parametric NLP problem:

\begin{equation}
\begin{gathered}
\label{equation: 12}
\begin{array}{l}
	\underset{\boldsymbol{z}}{\min}\,\,\phi \left( \boldsymbol{z};\boldsymbol{\theta }^0 \right)\\
	s.t.\;\boldsymbol{c}\left( \boldsymbol{z};\boldsymbol{\theta }^0 \right) =0\\
	\;\;\;\;\;\;\boldsymbol{z}\geqslant 0\\
\end{array}
\end{gathered}
\end{equation}
\noindent where the variable $\boldsymbol{z}$ consists of discretized $\bar{\boldsymbol{x}}^i$ and $\bar{u}^i$.

The NLP problem (\ref{equation: 12}) is solved by the interior point method (IPM) with bound constraints $\boldsymbol{z}\geqslant 0$ replaced by logarithmic barrier terms that are added to the objective function. Thus, problem (\ref{equation: 12}) is reformulated as:

\begin{equation}
\begin{gathered}
\label{equation: 13}
\begin{array}{l}
	\underset{\boldsymbol{z}}{\min}\,\,\phi \left( \boldsymbol{z};\boldsymbol{\theta }^0 \right) -\mu \sum{\ln \left( \boldsymbol{z} \right)}\\
	s.t.\;\boldsymbol{c}\left( \boldsymbol{z};\boldsymbol{\theta }^0 \right) =\boldsymbol{0}\\
\end{array}
\end{gathered}
\end{equation}
\noindent where the solution of problem (\ref{equation: 13}) becomes that of (\ref{equation: 12}) as the barrier parameter $\mu \rightarrow 0$. The Karush-Kuhn-Tucker (KKT) conditions for problem (\ref{equation: 13}) are defined as:

\begin{equation}
\begin{gathered}
\label{equation: 14}
\begin{array}{l}
  \underset{\boldsymbol{z}}{\min}\,\,\phi \left( \boldsymbol{z};\boldsymbol{\theta }^0 \right) -\mu \sum{\ln \left( \boldsymbol{z} \right)}\\
  s.t.\;\boldsymbol{c}\left( \boldsymbol{z};\boldsymbol{\theta }^0 \right) =\boldsymbol{0}\\
\end{array}
\end{gathered}
\end{equation}
\noindent where the Lagrange function is defined as $L=\phi \left( \boldsymbol{z};\boldsymbol{\theta }^0 \right) +\boldsymbol{c}\left( \boldsymbol{z};\boldsymbol{\theta }^0 \right) ^T\boldsymbol{\lambda }-\boldsymbol{z}^T\boldsymbol{\nu }$; $\boldsymbol{\nu}$ and $\boldsymbol{\lambda }$ are the KKT multipliers; $\boldsymbol{s}^*\left( \boldsymbol{\theta }^0 \right) =\left[ \begin{matrix}
	\boldsymbol{z}^{*T}&		\boldsymbol{\lambda }^T&		\boldsymbol{\nu}^T\\
\end{matrix} \right] ^T$ is the solution vector; $\boldsymbol{Z}^*=\mathrm{diag}\left( \boldsymbol{z}^* \right)$.

Expanding the KKT conditions at the optimal solution with parameter $\boldsymbol{\theta }\ne \boldsymbol{\theta }^0$ leads to:

\begin{equation}
\begin{gathered}
\label{equation: 15}
\boldsymbol{\varPsi }\left( \boldsymbol{s}^*\left( \boldsymbol{\theta } \right) \right) =\boldsymbol{\varPsi }\left( \boldsymbol{s}^*\left( \boldsymbol{\theta }^0 \right) \right) +\frac{d\boldsymbol{\varPsi }\left( \boldsymbol{s}^*\left( \boldsymbol{\theta }^0 \right) \right) ^{\mathrm{T}}}{d\boldsymbol{\theta }}\Delta \boldsymbol{\theta }+O\left( \left\| \Delta \boldsymbol{\theta } \right\| ^2 \right) 
\end{gathered}
\end{equation}
\noindent where $\boldsymbol{\varPsi }\left( \boldsymbol{s}^*\left( \boldsymbol{\theta }^0 \right) \right) =\boldsymbol{0}$ and $\Delta \boldsymbol{\theta }=\boldsymbol{\theta }-\boldsymbol{\theta }^0$. Utilizing the implicit function theorem, Equation (\ref{equation: 15}) is written as:

\begin{equation}
\begin{gathered}
\label{equation: 16}
\frac{d\boldsymbol{\varPsi }\left( \boldsymbol{s}^*\left( \boldsymbol{\theta }^0 \right) \right) ^T}{d\boldsymbol{\theta }}\Delta \boldsymbol{\theta }=\left( \nabla _{\boldsymbol{s}}\boldsymbol{\varPsi }^T\frac{d\boldsymbol{s}}{d\boldsymbol{\theta }}+\nabla _{\boldsymbol{\theta }}\boldsymbol{\varPsi } \right) ^T\Delta \boldsymbol{\theta }\approx \boldsymbol{0}
\end{gathered}
\end{equation}
\noindent where $\boldsymbol{M}=\nabla _{\boldsymbol{s}}\boldsymbol{\varPsi }^T=\left[ \begin{matrix}
	\nabla _{\boldsymbol{zz}}L\left( \boldsymbol{s}^*\left( \boldsymbol{\theta }^0 \right) \right)&		\nabla _{\boldsymbol{z}}\boldsymbol{c}\left( \boldsymbol{z}^*\left( \boldsymbol{\theta }^0 \right) \right)&		-\boldsymbol{I}\\
	\nabla _{\boldsymbol{z}}\boldsymbol{c}\left( \boldsymbol{z}^*\left( \boldsymbol{\theta }^0 \right) \right) ^T&		\mathbf{0}&		\mathbf{0}\\
	\boldsymbol{\varGamma }\left( \boldsymbol{\theta }^0 \right)&		\mathbf{0}&		\boldsymbol{Z}^*\left( \boldsymbol{\theta }^0 \right)\\
\end{matrix} \right]$ is the KKT matrix, which is directly available in the factored form from the solution to problem (\ref{equation: 13}); $\boldsymbol{\varGamma }=\mathrm{diag}\left( \boldsymbol{\nu} \right)$ and $\boldsymbol{N}_{\boldsymbol{\theta }}=\nabla _{\boldsymbol{\theta }}\boldsymbol{\varPsi }^T=\left[ \begin{matrix}
	\nabla _{\boldsymbol{z}\theta}L\left( \boldsymbol{s}^*\left( \boldsymbol{\theta }^0 \right) \right)&		\nabla _{\boldsymbol{\theta }}\boldsymbol{c}\left( \boldsymbol{z}^*\left( \boldsymbol{\theta }^0 \right) \right) ^T&		0\\
\end{matrix} \right] ^T$. The KKT matrix $\boldsymbol{M}$ is directly available in the factored form from the solution to problem (\ref{equation: 13}). This is because solving an NLP problem in the IPM solver (e.g. IPOPT) is replaced by an approximation through solving a linear system, where the left-hand side is the KKT matrix $\boldsymbol{M}$ \cite{[26]wachter2006implementation}.

Thus, for the perturbation problem with $\boldsymbol{\theta }^k$, its solution can be approximated by:

\begin{equation}
\begin{gathered}
\label{equation: 17}
\begin{array}{c}
	\boldsymbol{M}\Delta \boldsymbol{s}_{\boldsymbol{\theta }}\approx -\boldsymbol{N}_{\boldsymbol{\theta }}\left( \boldsymbol{\theta }^k-\boldsymbol{\theta }^0 \right)\\
	\boldsymbol{s}\left( \boldsymbol{\theta }^k \right) =\boldsymbol{s}\left( \boldsymbol{\theta }^0 \right) +\Delta \boldsymbol{s}_{\boldsymbol{\theta }}\\
\end{array}
\end{gathered}
\end{equation}

Through Equation (\ref{equation: 17}), $R^i\left( \boldsymbol{\theta }^k \right)$ is rapidly obtained through NLP sensitivity analysis calculated by each parafoil. Consequently, the landing point allocation can be determined by minimizing the overall control energy consumption, which is formulated as the following assignment problem:

\begin{equation}
\begin{gathered}
\label{equation: 18}
\begin{array}{l}
	\underset{a^{ik}}{\min}\sum\nolimits_{i=1}^N{\sum\nolimits_{k=1}^N{R^i\left( \boldsymbol{\theta }^k \right)}}a^{ik}\\
	s.t.\,\sum\nolimits_{i=1}^N{a^{ik}=1},k=1,..,N\\
	\quad \,\,\,\,\sum\nolimits_{k=1}^N{a^{ik}=1},i=1,..,N\\
	\quad \,\,\,\,a^{ik}=\begin{cases}
	1,i\mathrm{th}\,\mathrm{parafoil}\,\mathrm{to}\,k\mathrm{th}\,\mathrm{landing}\,\mathrm{point}\\
	0,\mathrm{others}\\
\end{cases}\\
\end{array}
\end{gathered}
\end{equation}
\noindent where $a^{ik}$ denotes the allocation decision variable; the constraints indicate that each landing point can only be allocated to one parafoil, and each parafoil can only touch down at one landing point.

The cost matrix and allocation matrix for the landing point allocation can be described as:

\begin{equation}
\begin{gathered}
\label{equation: 19}
\boldsymbol{R}= \begin{bmatrix}
	R^1\left( \boldsymbol{\theta }^1 \right)&		\cdots&		R^1\left( \boldsymbol{\theta }^N \right)\\
	\vdots&		\ddots&		\vdots\\
	R^N\left( \boldsymbol{\theta }^1 \right)&		\cdots&		R^N\left( \boldsymbol{\theta }^N \right)\\
\end{bmatrix}
\end{gathered}
\end{equation}

\begin{equation}
\begin{gathered}
\label{equation: 20}
\boldsymbol{A}=\begin{bmatrix}
	a^{11}&		\cdots&		a^{1N}\\
	\vdots&		\ddots&		\vdots\\
	a^{N1}&		\cdots&		a^{NN}\\
\end{bmatrix}
\end{gathered}
\end{equation}

The Hungarian algorithm \cite{[27]kuhn1955hungarian} is leveraged to solve the proposed assignment problem (\ref{equation: 18}). The algorithm begins by subtracting the smallest element from each row and then subtracting the smallest element from each column of the cost matrix. Then, the algorithm covers all zeros in the resulting matrix using a minimum number of horizontal and vertical lines. If $N$ lines are required, the optimal assignment is found at the position of all zeros. Otherwise, the algorithm finds the smallest element that is not covered by the line, subtracts it from all uncovered elements, adds it to all elements that are covered twice, and goes back to the previous line covering step. This iterative process continues until the optimal assignment is found. The Hungarian algorithm guarantees an optimal solution and solves the assignment problem in polynomial time, making it practical for the landing point allocation of multiple parafoil.

Through the terminal landing point allocation, the terminal constraints in Equation (\ref{equation: 5}) are converted to the following form:

\begin{equation}
\begin{gathered}
\label{equation: 21}
x^i\left( t_{f}^{i} \right) =x_{f}^{i},y^i\left( t_{f}^{i} \right) =y_{f}^{i},z^i\left( t_{f}^{i} \right) =z_f,i=1,...,N
\end{gathered}
\end{equation}
\noindent where $\left( x_{f}^{i},y_{f}^{i},z_f \right)$ is the allocated landing point for \textit{i}th parafoil, and the corresponding control energy is denoted as $R^i$. The trajectory of \textit{i}th parafoil to its corresponding landing point is denoted as $\boldsymbol{\chi }^i$, which is determined by solving problem (\ref{equation: 11}) initialized using the NLP sensitivity analysis result.

\subsection{Collision-free trajectory replanning algorithm}
After the terminal point allocation, the remaining coupled constraint is the collision avoidance constraint in Equation (\ref{equation: 4}). In order to decouple multiple parafoil landing trajectory optimization problem, we follow the idea of trajectory freezing \cite{[28]wang2023trajectory}, where the trajectories of other vehicles are considered fixed for each vehicle, allowing us to iteratively solve the decoupled problem.

To start with, we collect the trajectory $\boldsymbol{\chi }^i$ of each parafoil to detect the potential collision. Specifically, the relative distance between the parafoils at each discrete time point is calculated. If the distance is less than the threshold $d_s$, it is considered that a collision risk exists for the two parafoils. Through collision detection, a set containing parafoils that have collision trajectories is obtained, which is defined as follows:

\begin{equation}
\begin{gathered}
\label{equation: 22}
\mathbb{N} _c=\left\{ i|\forall j\in \mathbb{N} _c,\left\| \bar{\boldsymbol{r}}_{R}^{i}-\bar{\boldsymbol{r}}_{R}^{j} \right\| \leqslant d_{safe},i\ne j \right\} 
\end{gathered}
\end{equation}
\noindent where $\bar{\boldsymbol{r}}_{R}^{i}$ and $\bar{\boldsymbol{r}}_{R}^{j}$ are obtained from the reference trajectories $\boldsymbol{\chi }^i$ and $\boldsymbol{\chi }^j$, which omits the consideration of collision avoidance. The set $\mathbb{N} _c$ is sorted according to the control energy $R$ of its elements in ascending order to obtain $\mathbb{N} _{c,sort}$, which is:

\begin{equation}
\begin{gathered}
\label{equation: 23}
\mathbb{N} _{c,sort}=\left\{ b_1,...,b_c|b_1,...,b_c\in \mathbb{N} _c,R^{b_1}\leqslant ...\leqslant R^{b_c} \right\}  
\end{gathered}
\end{equation}

For each $b_i\in \mathbb{N} _{c,sort}$, the corresponding parafoil is chosen to replan the trajectory. The trajectories of other parafoils are considered fixed and leveraged to construct the inter-parafoil collision avoidance constraint. For example, we have the following collision avoidance constraint for parafoil $b_i$:

\begin{equation}
\begin{gathered}
\label{equation: 24}
\left\| \boldsymbol{r}_{R}^{b_i}-\bar{\boldsymbol{r}}_{R}^{s} \right\| ^2\geqslant \left( d_s \right) ^2,s=1,...,N,s\ne b_i
\end{gathered}
\end{equation}

Therefore, the multiple parafoil landing problem is decoupled to the following subproblem for each parafoil system in set $\mathbb{N} _{c,sort}$:

\begin{equation}
\begin{gathered}
\label{equation: 25}
\begin{array}{l}
	\underset{\boldsymbol{u}^{b_i},t_{f}^{b_i}}{\min}\,\,\int_0^{t_{f}^{b_i}}{\left\| \boldsymbol{u}^{b_i}\left( t \right) \right\| ^2dt}+\beta \sum{\left( \boldsymbol{p}+\boldsymbol{n} \right) ^{\mathrm{T}}}\boldsymbol{e}\\
	s.t.\,\dot{\boldsymbol{x}}^{b_i}\left( t \right) =f^{b_i}\left( \boldsymbol{x}^{b_i}\left( t \right) ,\boldsymbol{u}^{b_i}\left( t \right) \right)\\
	\quad \,\,\,\,0\leqslant \delta _{s}^{b_i}\left( t \right) ,\delta _{s}^{b_i}\left( t \right) +\delta _{a}^{b_i}\left( t \right) \leqslant \delta _{\max},\delta _{s}^{b_i}\left( t \right) -\delta _{a}^{b_i}\left( t \right) \leqslant \delta _{\max}\\
	\quad \,\,\,\,\left\| \boldsymbol{r}_{R}^{b_i}\left( t \right) -\bar{\boldsymbol{r}}_{R}^{s}\left( t \right) \right\| ^2\geqslant \left( d_s \right) ^2,s=1,...,N,s\ne b_i\\
	\quad \,\,\,\,\boldsymbol{x}^{b_i}\left( 0 \right) =\boldsymbol{x}_{init}^{b_i}\\
	\quad \,\,\,\,x^{b_i}\left( t_{f}^{b_i} \right) -\theta _{x}^{b_i}=p_x-n_x\\
	\quad \,\,\,\,y^{b_i}\left( t_{f}^{b_i} \right) -\theta _{y}^{b_i}=p_y-n_y\\
	\quad \,\,\,\,z^{b_i}\left( t_{f}^{b_i} \right) -z_f=p_z-n_z\\
	\quad \,\,\,\,\psi ^{b_i}\left( t_{f}^{b_i} \right) -\alpha _{anti\_wind}=p_{\psi}-n_{\psi}\\
	\quad \,\,\,\,\boldsymbol{p}\geqslant 0,\boldsymbol{n}\geqslant 0\\
\end{array}
\end{gathered}
\end{equation}

In problem (\ref{equation: 25}), collision avoidance is strictly enforced as the safety of the parafoils is the priority. Meanwhile, the terminal landing requirement is relaxed by introducing the penalty term, which reduces the difficulty of the solving process. In other words, when it becomes challenging to satisfy both the collision avoidance and terminal landing requirements, terminal landing accuracy is compromised to avoid the collision of parafoils. Through solving problem (\ref{equation: 25}), the trajectory $\boldsymbol{\chi }^{b_i}$ of parafoil $b_i$ is updated. The subsequent process continues until all elements in $\mathbb{N} _{c,sort}$ is considered, which corresponds to all parafoils with potential collision trajectories have replanned non-collision trajectories.

The rationale behind replanning the trajectories of parafoil systems following ascending order of control energy consumption is twofold. One is that parafoils with less control energy consumption correspond to greater collision avoidance capability. Therefore, they are given higher priority to undertake more collision maneuvers. Another is that the simultaneous update of all vehicle trajectories may not converge, as indicated in \cite{[28]wang2023trajectory}. Hence, we sequentially replan the trajectory for each parafoil at risk of collision, ensuring the procedure converges under the circumstance that all sub-problems are successfully solved.

\subsection{NMPC trajectory tracking algorithm}
The NMPC algorithm is adopted to track the generated trajectory and provide the control command. Different from the conventional linear model predictive control (LMPC) for parafoil control, the NMPC utilizes the nonlinear dynamics of the parafoil system, which results in better trajectory tracking performance \cite{[23]wei2024dynamic}.

The collision-free reference trajectory for \textit{i}th parafoil contains the state profile $\boldsymbol{x}_{ref}^{i}$ and control profile $\boldsymbol{u}_{ref}^{i}$, which is to be tracked by the NMPC controller during each guidance horizon. Hence, the following objective is minimized to find the optimal control command.

\begin{equation}
\begin{gathered}
\label{equation: 27}
\begin{array}{l}
	J=\sum_{l=0}^{N_{track}-1}{\left( \begin{array}{c}
	\left\| \boldsymbol{x}_{l}^{i}-\boldsymbol{x}_{ref}^{i}\left( t_n+lT_{track} \right) \right\| _{\boldsymbol{Q}}^{2}\\
	+\left\| \boldsymbol{u}_{l}^{i}-\boldsymbol{u}_{ref}^{i}\left( t_n+lT_{track} \right) \right\| _{\boldsymbol{R}}^{2}\\
\end{array} \right)}\\
	\quad \,\,\,\,+\left\| \boldsymbol{x}_{N_{track}}^{i}-\boldsymbol{x}_{ref}^{i}\left( t_n+N_{track}T_{track} \right) \right\| _{\boldsymbol{Q}_f}^{2}\\
\end{array}
\end{gathered}
\end{equation}
\noindent where $\boldsymbol{Q}$ and $\boldsymbol{R}$ are symmetric positive definite matrices; $N_{track}$ is the prediction horizon length; $T_{track}$ is the tracking time interval; The NMPC is to determine the optimal control $\boldsymbol{u}^*\left( t \right)$ that minimizes the objective in Equation (\ref{equation: 27}) and satisfies the following constraints during each guidance horizon.

\begin{equation}
\begin{gathered}
\label{equation: 28}
\begin{array}{l}
	\boldsymbol{x}_{l+1}^{i}=f^i\left( \boldsymbol{x}_{l}^{i},\boldsymbol{u}_{l}^{i} \right)
\end{array}
\end{gathered}
\end{equation}

\begin{equation}
\begin{gathered}
\label{equation: 29}
\begin{array}{l}
	\boldsymbol{x}_{0}^{i}=\boldsymbol{x}^i\left( t_n \right)
\end{array}
\end{gathered}
\end{equation}

\begin{equation}
\begin{gathered}
\label{equation: 30}
\begin{array}{l}
	\begin{cases}
		0\leqslant \delta _{s,l}^{i}\\
		\delta _{s,l}^{i}+\delta _{a,l}^{i}\leqslant 1\\
		\delta _{s,l}^{i}-\delta _{a,l}^{i}\leqslant 1\\
	\end{cases}
\end{array}
\end{gathered}
\end{equation}
\noindent where the constraint in Equation (\ref{equation: 28}) is the discretized parafoil system dynamics via the simultaneous collocation method \cite{[25]kameswaran2006simultaneous}, Equation (\ref{equation: 29}) is the updated initial state, and Equation (\ref{equation: 30}) is the limits of control variables. As such, the optimal trajectory tracking guidance problem for NMPC is formulated as follows:

\begin{equation}
\begin{gathered}
\label{equation: 31}
\begin{array}{l}
	\underset{\boldsymbol{x}^*,\boldsymbol{u}^*}{\min}\sum_{l=0}^{N_{track}-1}{\left( \begin{array}{c}
	\left\| \boldsymbol{x}_{l}^{i}-\boldsymbol{x}_{ref}^{i}\left( t_n+lT_{track} \right) \right\| _{\boldsymbol{Q}}^{2}\\
	+\left\| \boldsymbol{u}_{l}^{i}-\boldsymbol{u}_{ref}^{i}\left( t_n+lT_{track} \right) \right\| _{\boldsymbol{R}}^{2}\\
\end{array} \right)}\\
	\quad \quad +\left\| \boldsymbol{x}_{N_{track}}^{i}-\boldsymbol{x}_{ref}^{i}\left( t_n+N_{track}T_{track} \right) \right\| _{\boldsymbol{Q}_f}^{2}\\
	s.t.\;\;\boldsymbol{x}_{l+1}^{i}=f^i\left( \boldsymbol{x}_{l}^{i},\boldsymbol{u}_{l}^{i} \right)\\
	\;\;\;\;\;\;\;\boldsymbol{x}_{0}^{i}=\boldsymbol{x}^i\left( t_n \right)\\
	\;\;\;\;\;\;\;0\leqslant \delta _{s,l}^{i},\delta _{s,l}^{i}+\delta _{a,l}^{i}\leqslant 1,\delta _{s,l}^{i}-\delta _{a,l}^{i}\leqslant 1\\
	\;\;\;\;\;\;\;l=0,...,N_{track}-1\\
\end{array}
\end{gathered}
\end{equation}

At time $t_n - T_{track}$, the initial state $\boldsymbol{x}^i\left( t_n \right)$ is predicted and the NMPC problem (\ref{equation: 31}) is solved to generate the optimal trajectory tracking command $\boldsymbol{u}^{i,*}\left( t_n+l T_{track} \right) ,l=0,...,N_{track}$. At time $t_n$, the control command $u^{i,*}\left( t_n \right)$ is utilized to control the \textit{i}th parafoil.

\subsection{Model simplification and moving horizon correction}
Due to the highly nonlinear characteristics of the parafoil dynamics \cite{[A1]sun2020trajectory} and its computational complexity \cite{[23]wei2024dynamic}, solving trajectory optimization problem (\ref{equation: 11}) and problem (\ref{equation: 25}) can be time-consuming and impractical for the online update of the collision-free trajectories.

In order to boost the computation speed, the dynamic model in problems (\ref{equation: 11})(\ref{equation: 25}) is replaced by the simplified kinematic model as follows:

\begin{equation}
\begin{gathered}
\label{equation: 10}
\begin{array}{l}
	\left[ \begin{array}{c}
	\dot{x}^i\\
	\dot{y}^i\\
	\dot{z}^i\\
	\dot{\psi}^i\\
\end{array} \right] =\left[ \begin{array}{c}
	v_{h}^{i}\cos \left( \psi ^i \right) +w_x\\
	v_{h}^{i}\sin \left( \psi ^i \right)\\
	-v_{d}^{i}\\
	u_{\psi}^{i}\\
\end{array} \right]\\
	-\dot{\psi}_{\max}^{i}\leqslant u_{\psi}^{i}\leqslant \dot{\psi}_{\max}^{i}\\
\end{array}
\end{gathered}
\end{equation}
\noindent where $v_{h}^{i}$ and $v_{d}^{i}$ are the estimated constant horizontal velocity and vertical velocity; $w_x$ is the magnitude of the known constant wind; $\dot{\psi}_{\max}^{i}$ is the maximum turning rate; the kinematic model is denoted as $\dot{\hat{\boldsymbol{x}}}^i\left( t \right) =\hat{\boldsymbol{f}}^i\left( \hat{\boldsymbol{x}}^i\left( t \right) ,\hat{u}^i\left( t \right) \right)$, where $\hat{\boldsymbol{x}}^i\left( t \right) =\left[ \begin{matrix}
	x^i&		y^i&		z^i&		\psi ^i\\
\end{matrix} \right] ^{\mathrm{T}}$ and $\hat{u}^i\left( t \right) =u_{\psi}^{i}$.

In practice, the steady-state gliding condition \cite{[16]ward2013adaptive} assumed by the parafoil kinematic model does not hold due to nonlinear dynamics, consistent control input, and aerodynamics caused by the wind field. This results in the mismatch between the kinematic model and the actual parafoil system, which not only deteriorates the landing accuracy but also might fail to avoid collision. To address this issue, a moving horizon correction method is developed to update the kinematic model.

For the moving horizon correction process, at time $t_n$, the latest $N_{real}+1$ measured partial states $\hat{\boldsymbol{x}}_{real}^{i}\left( t_n+\left( l-N_{real} \right) T_{track} \right) =\left[ \begin{matrix}
	x_{l}^{i}&		y_{l}^{i}&		z_{l}^{i}&		\psi _{l}^{i}\\
\end{matrix} \right] ^{\mathrm{T}}$ obtained at the sampling time instances $l=0,...,N_{real}$ are utilized. The model correction problem can be formulated as follows:

\begin{equation}
\begin{gathered}
\label{equation: 26}
\begin{array}{l}
	\underset{\boldsymbol{\lambda }_{cor}^{i}}{\min}\sum_{k=1}^{N_{real}}{\left( \begin{array}{c}
	\left\| \hat{\boldsymbol{x}}_{l}^{i}-\hat{\boldsymbol{x}}_{real}^{i}\left( t_n+\left( l-N_{real} \right) T_{track} \right) \right\| _{\boldsymbol{Q}_{cor}}^{2}\\
	+\left\| \boldsymbol{\lambda }_{cor}^{i} \right\| _{\boldsymbol{R}_{cor}}^{2}\\
\end{array} \right)}\\
	s.t.\;\;\dot{\hat{\boldsymbol{x}}}_{l+1}^{i}=\hat{f}^i\left( \hat{\boldsymbol{x}}_{l}^{i},\hat{u}_{l}^{i},\boldsymbol{\lambda }_{cor}^{i} \right)\\
	\;\;\;\;\;\;\;\hat{\boldsymbol{x}}_{0}^{i}=\boldsymbol{Cx}_{real}^{i}\left( t_n-N_{real}T_{track} \right)\\
	\;\;\;\;\;\;\;l=0,...,N_{real}-1\\
\end{array}
\end{gathered}
\end{equation}
\noindent where $\boldsymbol{C}$ is the binary diagonal matrix, which corresponds to state variables shared by dynamics and kinematics; $\boldsymbol{Q}_{cor}$ and $\boldsymbol{R}_{cor}$ are symmetric positive definite matrices; $N_{real}$ is the moving horizon length; $\boldsymbol{\lambda }_{cor}^{i}$ is the model correction parameter to be optimized. Usually, an initial state estimation term should be introduced in the objective function to account for the state estimation error. However, we assume the initial state of the vehicle is known, as in \cite{[29]chai2019integrated}.

Inspired by the steady-state gliding condition of the parafoil system, we account for model inaccuracies by modifying the constant horizontal and vertical velocities in the kinematic model, making the modified kinematic model close to the actual system. Hence, the model correction parameter is chosen as $\boldsymbol{\lambda }_{cor}^{i}=\left[ \begin{matrix}
	\lambda _{h}^{i}&		\lambda _{d}^{i}\\
\end{matrix} \right] ^{\mathrm{T}}$, and the corrected velocities are $\hat{v}_{h}^{i}=v_{h}^{i}+\lambda _{h}^{i}$ and $\hat{v}_{d}^{i}=v_{d}^{i}+\lambda _{d}^{i}$ for each parafoil system.

The moving horizon mechanism is preferred because it only takes into account the latest $N_{real}+1$ measurements, which reduces the computational complexity by restricting the dimension of nonlinear optimization problems \cite{[30]rao2003constrained}. The proposed moving horizon correction method preserves the simplicity of the kinematic model to reduce the computational burden while being efficient in correcting the model mismatch.

\subsection{Overall framework}
\begin{figure}[!ht]
\centering
\includegraphics[width=2.7in]{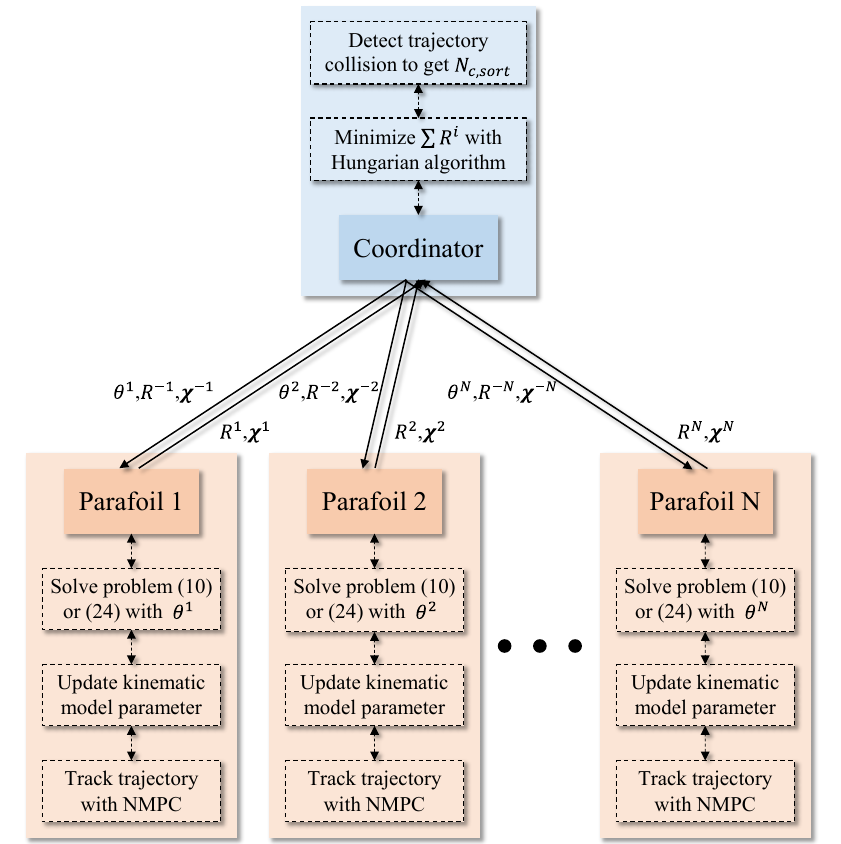}\\
\caption{Illustration of the proposed approach.}
\label{Fig2}
\end{figure}

\begin{figure}[!ht]
\centering
\includegraphics[width=3.4in]{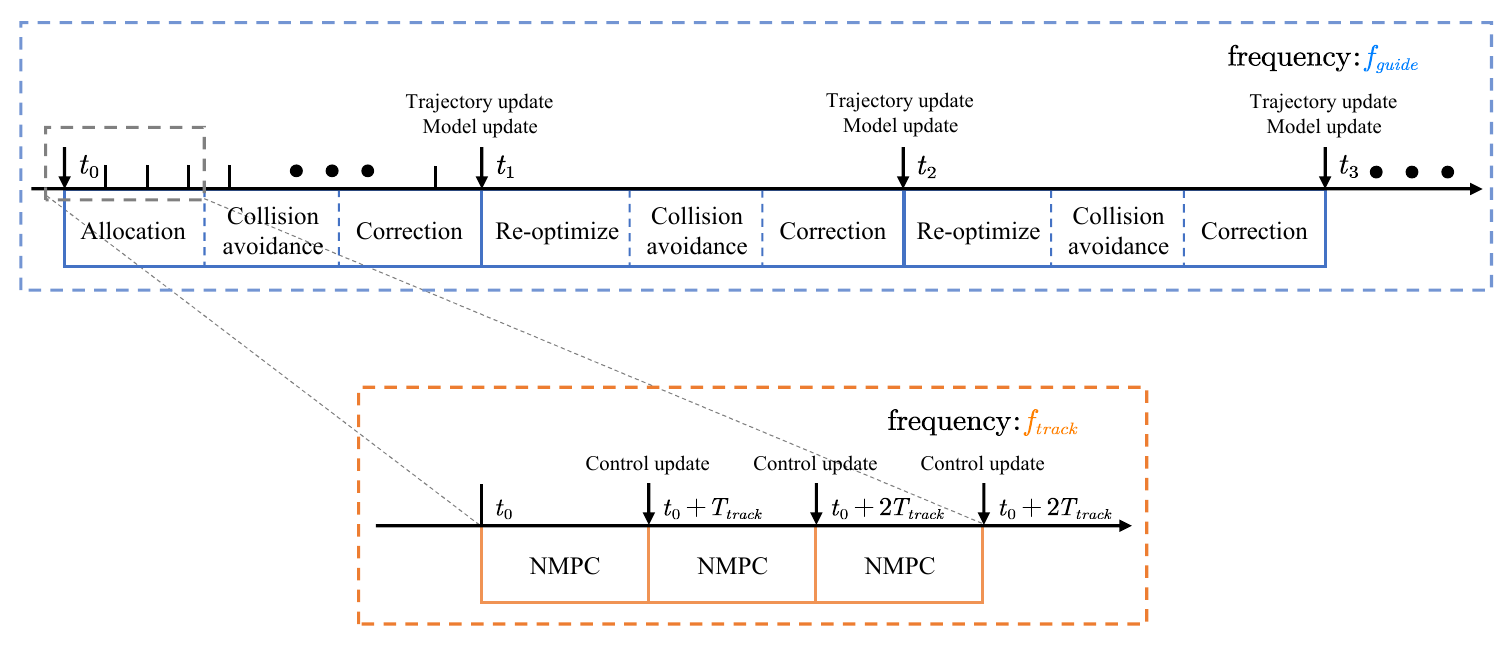}\\
\caption{Timeline of the proposed approach calculation.}
\label{Fig3}
\end{figure}

The coordinated guidance and control approach is summarized in the pseudocode as Algorithm 1 (for the coordinator) and Algorithm 2 (for each parafoil). To better explain the algorithm, the structure is illustrated in Fig. \ref{Fig2} and the algorithm timeline is shown in Fig \ref{Fig3}. At time $t_0$, the landing point allocation algorithm is executed, and the collision avoidance trajectories are replanned if potential collision is detected. It is important to point out that the trajectory optimization problems are solved in parallel due to the decoupled problem formulation, and the trajectory to each landing point is rapidly computed via NLP sensitivity analysis. Then, the trajectories are updated with a frequency of $f_{guide}$ Hz. To clarify, at time $t_k,k>0$, the landing trajectories are re-optimized, the collision avoidance trajectories are replanned if potential collision is detected, and the kinematic model parameter $\boldsymbol{\lambda }_{cor}$ is updated. Within the guidance horizon $\left[ t_k,t_{k+1} \right]$, the NMPC algorithm is performed with a frequency of $f_{track}=1/T_{track}$ Hz. To be specific, at time $t_n=t_k+nT_{track},n=0,...,\lfloor f_{track}/f_{guide} \rfloor$, the optimal trajectory tracking command is generated the NMPC controller to track the reference trajectory separately in parafoil until time $t_{k+1}$. The subsequent process continues in a similar manner until all parafoils touch down.

For the proposed coordinated guidance and control approach, the original coupled trajectory optimization problem (\ref{equation: 8}) is decomposed into small decoupled problems computed separately in each parafoil. The number of sub-problem variables is unrelated to the parafoils’ number $N$, and the number of collision avoidance constraints is, to some extent, linearly related to $N$. In contrast, for the coupled trajectory optimization problem (\ref{equation: 8}), its number of variables is linearly related to $N$ and its number of collision avoidance constraints increases quadratically with $N$. Therefore, the proposed approach significantly decreases the problem size, which consequently reduces the computational complexity.

\begin{breakablealgorithm}
\caption{Coordinated guidance and control method \\ (for the coordinator).}
\begin{algorithmic}[1]
\STATE Receive NLP sensitivity analysis results from all parafoils;
\STATE Allocate landing points by solving problem (\ref{equation: 18});
\FOR {each guidance horizon}
	\STATE Receive trajectories from all parafoils, detect collision;
	\IF {trajectory collision is detected}
		\STATE Find set $N_{c,sort}$ defined by Equation (\ref{equation: 23});
		\FOR {each $b_i\in \mathbb{N} _{c,sort}$}
			\STATE Send other parafoil trajectories to parafoil $b_i$;
			\STATE Receive and update trajectory $\boldsymbol{\chi }^{b_i}$ from parafoil $b_i$;
		\ENDFOR
	\ENDIF
	\IF {all parafoils touch down}
		\STATE Terminate
	\ENDIF
\ENDFOR
\end{algorithmic}
\label{algorithm1}
\end{breakablealgorithm}

\begin{breakablealgorithm}
\caption{Coordinated guidance and control method \\ (for \textit{i}th parafoil).}
\begin{algorithmic}[1]
\STATE Solve problem (\ref{equation: 11}) and use NLP sensitivity analysis to evaluate each landing point, send results to coordinator;
\STATE Receive designated landing point $\theta ^i$ from coordinator, and calculate the corresponding reference trajectory $\boldsymbol{\chi }^i$;
\FOR {each guidance horizon}
	\STATE Update reference trajectory $\boldsymbol{\chi }^i$ by solving problem (\ref{equation: 11});
	\IF {replanning is required from coordinator}
		\STATE Receive other parafoils’ trajectories $\boldsymbol{\chi }^{-i}$ from coordinator, and update collision-free trajectory $\boldsymbol{\chi }^i$ by solving problem (\ref{equation: 25});
	\ENDIF
	\STATE Update the kinematic model parameters by solving problem (\ref{equation: 26});
	\FOR {each control horizon}
		\STATE Apply tracking command $\boldsymbol{u}^*$ with one step;
		\STATE Estimate next prediction state, and solve problem (\ref{equation: 31}) to obtain next tracking command $\boldsymbol{u}^*$;
		\IF {the parafoil touches down}
			\STATE Terminate
		\ELSE
			\STATE Detect the actual state of the parafoil;
		\ENDIF
	\ENDFOR
\ENDFOR
\end{algorithmic}
\label{algorithm2}
\end{breakablealgorithm}

\section{Results and simulation}
\label{Sec4}
\subsection{Simulation setup}
In this section, numerical results are presented to demonstrate the proposed coordinated guidance and control approach for multiple parafoil landing. Numerical simulations were performed in the Python environment and executed on a laptop with an i7-10750H CPU and 16GB RAM. Optimization problems were modeled through Pyomo \cite{[31]nicholson2018pyomo}, discretized by the simultaneous collocation method with Radau points \cite{[25]kameswaran2006simultaneous}, and solved by the open-source NLP solver IPOPT \cite{[26]wachter2006implementation}. The NLP sensitivity analysis was performed by the sIPOPT \cite{[32]pirnay2012optimal}.

\begin{figure}[!ht]
\centering
\includegraphics[width=15pc]{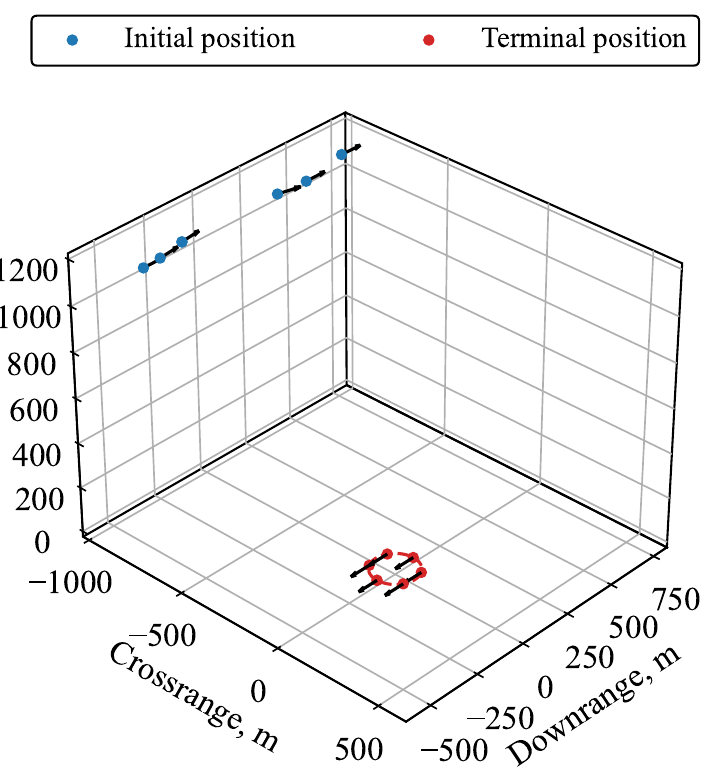}\\
\caption{Illustration of initial positions and landing points.}
\label{Fig4}
\end{figure}

\begin{figure}[!ht]
\centering
\includegraphics[width=15pc]{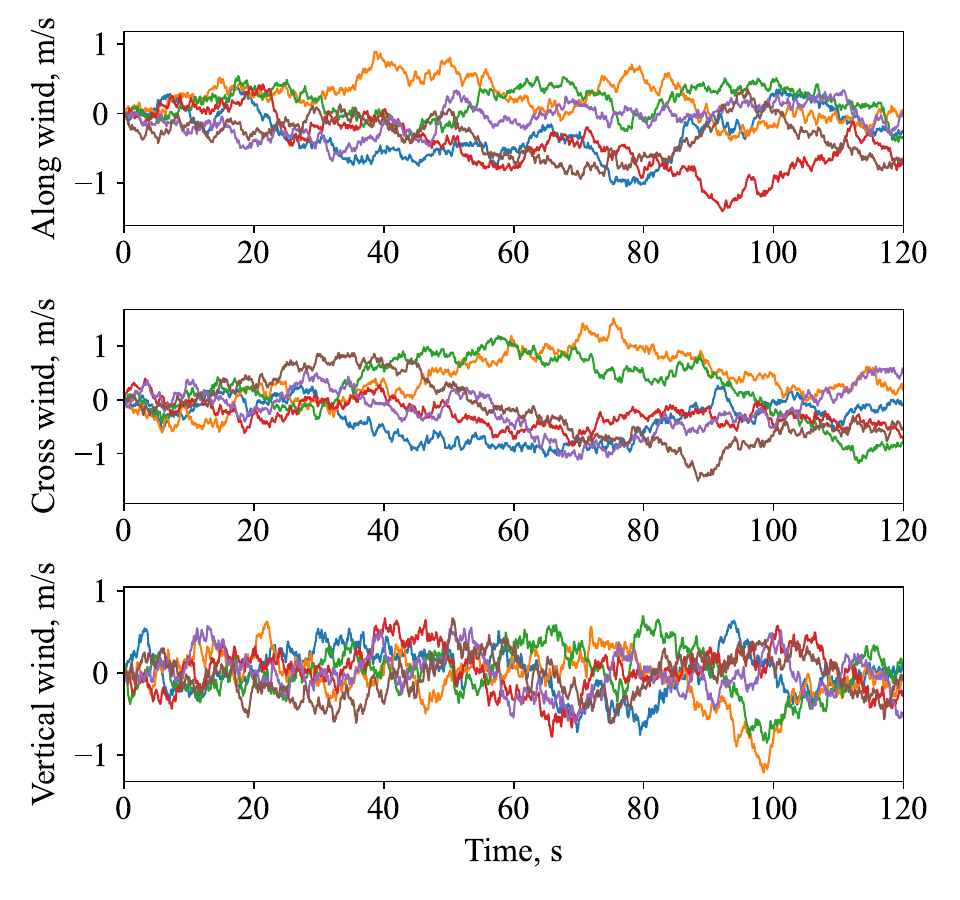}\\
\caption{Wind disturbances on each parafoil.}
\label{Fig5}
\end{figure}

The parameters of the parafoil system are referred to in \cite{[23]wei2024dynamic}. The aerodynamic coefficients of each parafoil are perturbed following a Gaussian distribution, with 5\% of the mean value as three standard deviations. The initial conditions of the parafoils are listed in Table \ref{Tab1}. The safety distance is $d_s=50\mathrm{m}$, and the landing area radius is $d_f=300\mathrm{m}$. The terminal landing points are decided as evenly distributed points on a circle with a radius of 100m. Fig. \ref{Fig4} illustrates the initial position of parafoils and the landing points in 3-D space. The wind field consists of the known constant wind $\boldsymbol{w}_c=3\mathrm{m}/\mathrm{s}$ and unknown disturbance wind $\boldsymbol{w}_{d}^{i}$ on each parafoil system. The disturbance wind is constructed as the Dryden wind following the US MIL-HDBK-197 standard, which is depicted in Fig. \ref{Fig5}. Taking the solution time of the NLP problems into consideration, the update frequencies of the coordinated guidance trajectory and the NMPC control command are set to $f_{guide}=0.1\mathrm{Hz}$ and $f_{track}=1\mathrm{Hz}$, respectively.

\begin{table}[!ht]
\caption{Initial conditions of the parafoils.}
\label{Tab1}
\centering
\begin{tabular}{cc}
\hline
Number & Initial condition\\
\hline
1 & $\begin{array}{l}
	\boldsymbol{r}_{R,init}^{1}=\left( -264.00,-818.96,1201.99 \right) \mathrm{m}\\
	\boldsymbol{o}_{B\gets R,init}^{1}=\left( -0.65,-0.80,-6.79 \right) \mathrm{deg}\\
	\boldsymbol{v}_{B,init}^{1}=\left( 15.06,0.24,7.87 \right) \mathrm{m}/\mathrm{s}\\
	\boldsymbol{\omega }_{B,init}^{1}=\left( -0.47,0.24,0.0583 \right) \mathrm{deg}/\mathrm{s}\\
\end{array}$\\
\hline
2 & $\begin{array}{l}
	\boldsymbol{r}_{R,init}^{2}=\left( -505.42,-769.46,1217.38 \right) \mathrm{m}\\
	\boldsymbol{o}_{B\gets R,init}^{2}=\left( 2.85,3.54,-0.23 \right) \mathrm{deg}\\
	\boldsymbol{v}_{B,init}^{2}=\left( 16.34,-0.84,8.28 \right) \mathrm{m}/\mathrm{s}\\
	\boldsymbol{\omega }_{B,init}^{2}=\left( -0.87,-0.14,-0.81 \right) \mathrm{deg}/\mathrm{s}\\
\end{array}$\\
\hline
3 & $\begin{array}{l}
	\boldsymbol{r}_{R,init}^{3}=\left( -452.84,-736.13,1249.04 \right) \mathrm{m}\\
	\boldsymbol{o}_{B\gets R,init}^{3}=\left( -3.93,-2.80,-6.01 \right) \mathrm{deg}\\
	\boldsymbol{v}_{B,init}^{3}=\left( 15.59,-0.60,8.55 \right) \mathrm{m}/\mathrm{s}\\
	\boldsymbol{\omega }_{B,init}^{3}=\left( -0.0339,0.0105,-0.226 \right) \mathrm{deg}/\mathrm{s}\\
\end{array}$\\
\hline
4 & $\begin{array}{l}
	\boldsymbol{r}_{R,init}^{4}=\left( 273.63,-706.80,1276.49 \right) \mathrm{m}\\
	\boldsymbol{o}_{B\gets R,init}^{4}=\left( 2.01,4.65,3.34\cdot 10^{-4} \right) \mathrm{deg}\\
	\boldsymbol{v}_{B,init}^{4}=\left( 16.43,-0.32,8.40 \right) \mathrm{m}/\mathrm{s}\\
	\boldsymbol{\omega }_{B,init}^{4}=\left( -0.14,-0.13,0.55 \right) \mathrm{deg}/\mathrm{s}\\
\end{array}$\\
\hline
5 & $\begin{array}{l}
	\boldsymbol{r}_{R,init}^{5}=\left( 75.60.-661.85,1321.77 \right) \mathrm{m}\\
	\boldsymbol{o}_{B\gets R,init}^{5}=\left( -4.18,-1.34,14.03 \right) \mathrm{deg}\\
	\boldsymbol{v}_{B,init}^{5}=\left( 15.46,-0.95,7.76 \right) \mathrm{m}/\mathrm{s}\\
	\boldsymbol{\omega }_{B,init}^{5}=\left( -0.87,0.99,0.94 \right) \mathrm{deg}/\mathrm{s}\\
\end{array}$\\
\hline
6 & $\begin{array}{l}
	\boldsymbol{r}_{R,init}^{6}=\left( 400.26,-645.93,1360.60 \right) \mathrm{m}\\
	\boldsymbol{o}_{B\gets R,init}^{6}=\left( -3.31,-2.07,0.96 \right) \mathrm{deg}\\
	\boldsymbol{v}_{B,init}^{6}=\left( 15.36,-0.91,9.24 \right) \mathrm{m}/\mathrm{s}\\
	\boldsymbol{\omega }_{B,init}^{6}=\left( -0.12,8\cdot 10^{-3},-0.35 \right) \mathrm{deg}/\mathrm{s}\\
\end{array}$\\
\hline
\end{tabular}
\end{table}

\subsection{Performance of landing point allocation algorithm}
\begin{figure*}[!ht]
\begin{subfigure}[t]{.5\textwidth}
	\centering
	\includegraphics[width=14pc]{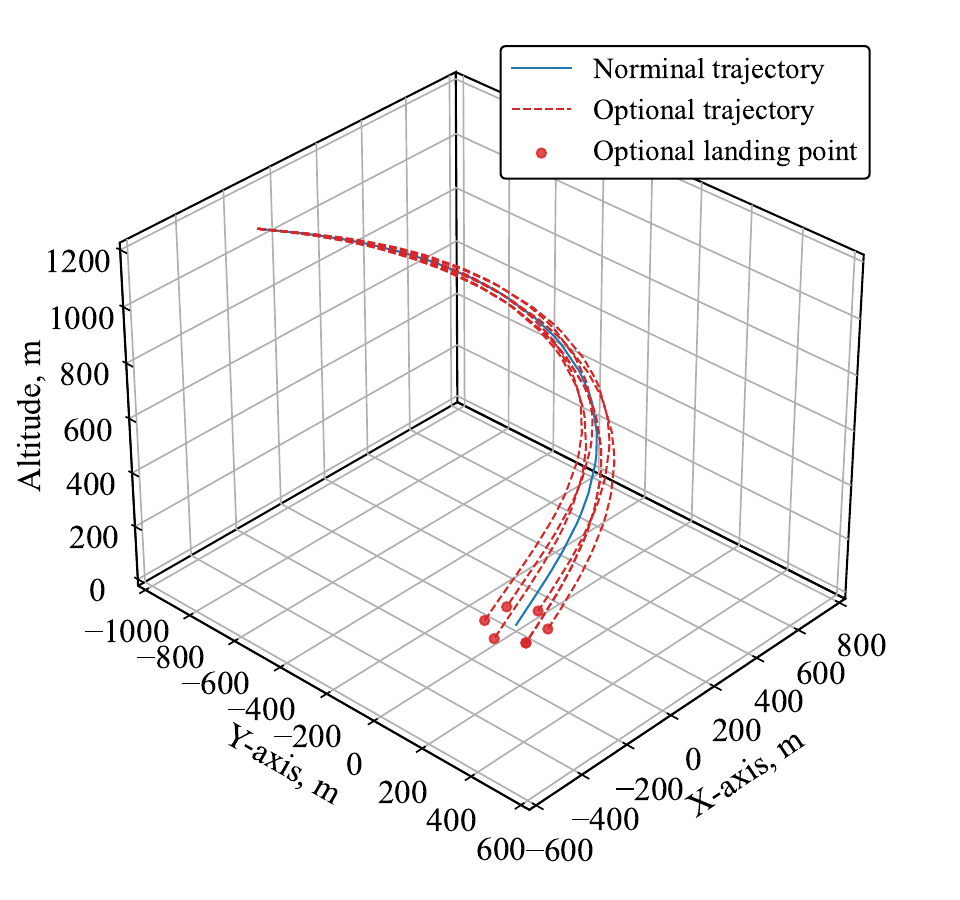}\\
	\caption{}
	\label{Fig6a}
\end{subfigure}
\begin{subfigure}[t]{.5\textwidth}
	\centering
	\includegraphics[width=16pc]{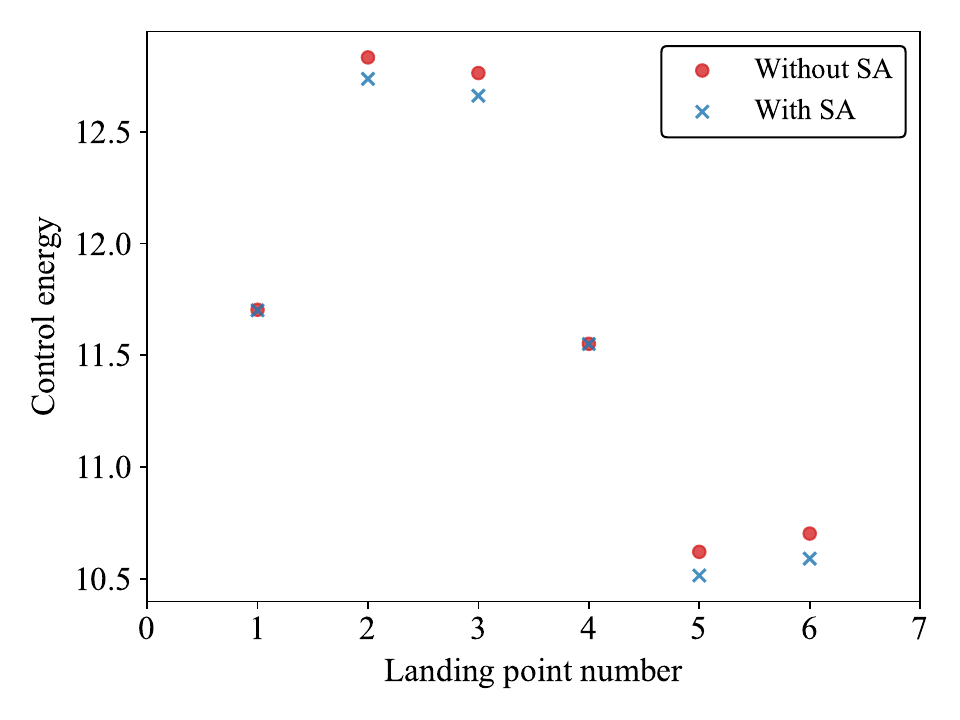}\\
	\caption{}
	\label{Fig6b}
\end{subfigure}
\caption{Landing point evaluation via NLP sensitivity analysis; (a) 3-D trajectory; (b) Control energy consumption.}
\label{Fig6}
\end{figure*}

\begin{figure}[!ht]
\centering
\includegraphics[width=15pc]{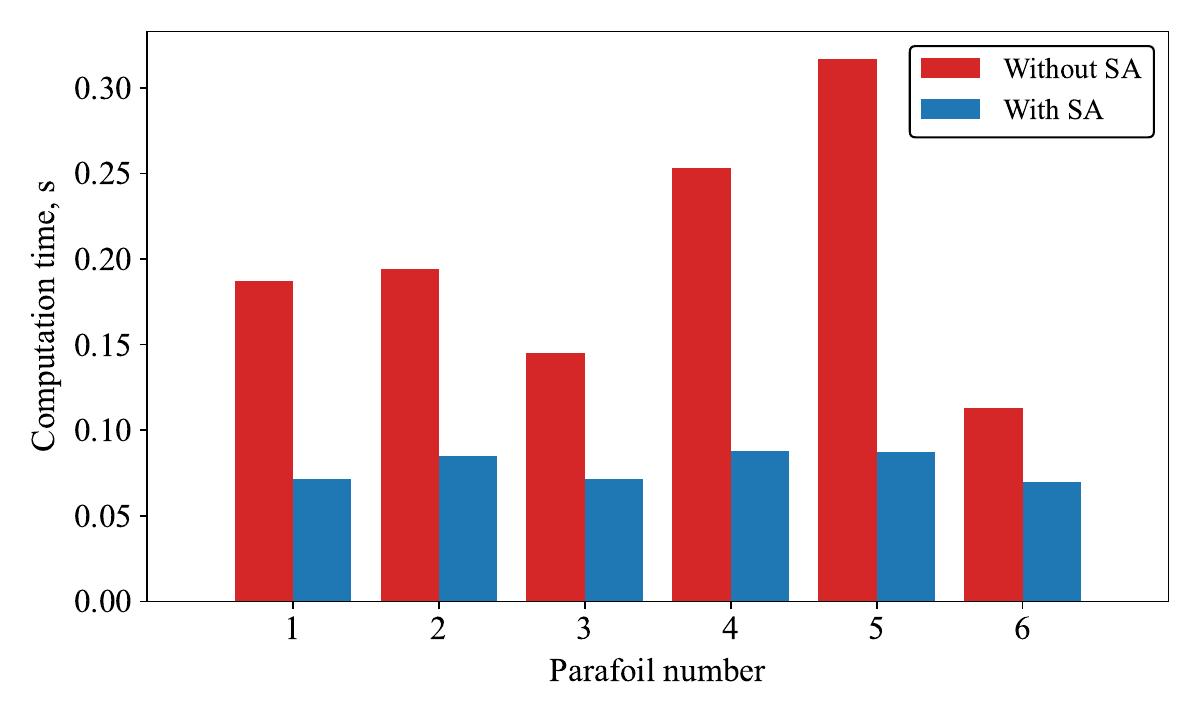}\\
\caption{Comparison of computation time.}
\label{Fig7}
\end{figure}

For each parafoil system, problem (\ref{equation: 11}) is solved to compute the trajectory to the landing area center. On this basis, the NLP sensitivity analysis is used to evaluate each landing point. An example of the parafoil is shown in Fig. \ref{Fig6}, where the control energy consumption to each landing point is efficiently estimated from NLP sensitivity analysis (SA). From Fig. \ref{Fig7}, the parafoils also consume less computation time than repeatedly solving the NLP problem.

Through the NLP sensitivity analysis, the performance indicator of each parafoil to each landing point is obtained, which forms the cost matrix $\boldsymbol{R}$ as shown in Table \ref{Tab2}. The Hungarian algorithm solves the corresponding assignment problem (\ref{equation: 18}) in 2.15 milliseconds and obtains the optimal landing point allocation, which is marked in blue.

\begin{table}[!ht]
\caption{Results of landing point allocation algorithm.}
\label{Tab2}
\centering
\begin{tabular}{ccccccc}
\hline
Parafoil & LP 1 & LP 2 & LP 3 & LP 4 & LP 5 & LP 6\\
\hline
1 & 11.70 & 12.74 & 12.66 & \cellcolor{NavyBlue!40} 11.55 & 10.52 & 10.59\\
2 & 11.47 & 12.59 & 12.54 & 11.36 & 10.24 & \cellcolor{NavyBlue!40} 10.29\\
3 & 12.52 & 13.62 & 13.53 & 12.34 & \cellcolor{NavyBlue!40} 11.24 & 11.33\\
4 & 9.78 & \cellcolor{NavyBlue!40} 10.61 & 10.67 & 9.91 & 9.08 & 9.02\\
5 & \cellcolor{NavyBlue!40} 10.95 & 12.22 & 12.56 & 11.63 & 10.35 & 10.02\\
6 & 11.73 & 12.65 & \cellcolor{NavyBlue!40} 12.68 & 11.81 & 10.89 & 10.86\\
\hline
\end{tabular}
\end{table}

\subsection{Performance of collision-free trajectory replanning algorithm}
\begin{figure*}[!ht]
\begin{subfigure}[t]{.5\textwidth}
	\centering
	\includegraphics[width=14pc]{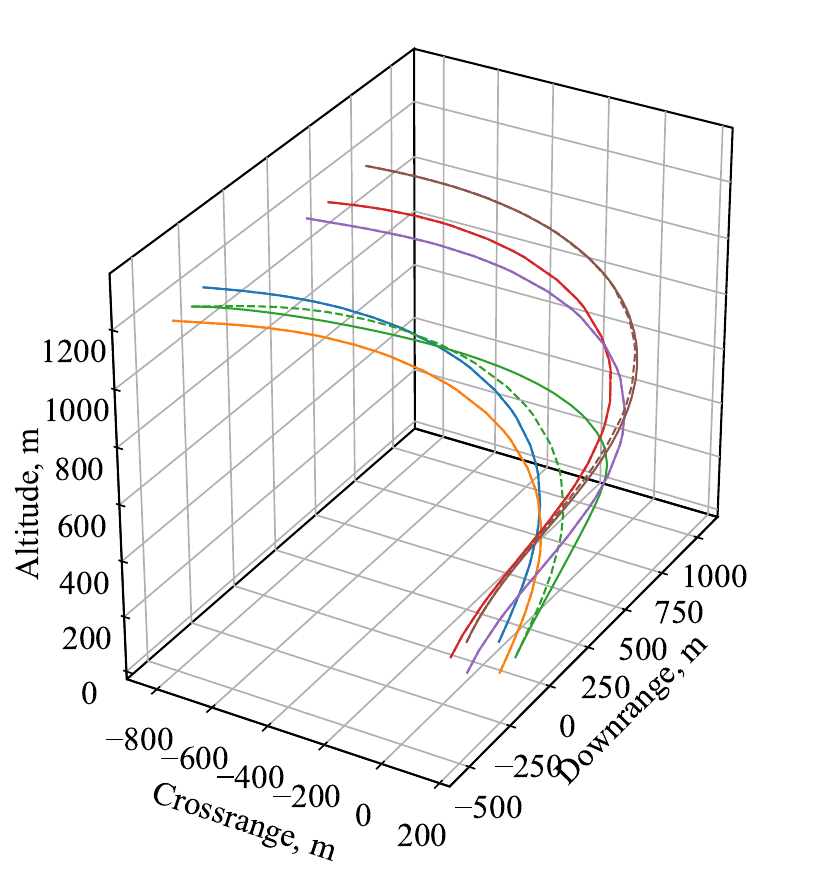}\\
	\caption{}
	\label{Fig8a}
\end{subfigure}
\begin{subfigure}[t]{.5\textwidth}
	\centering
	\includegraphics[width=16pc]{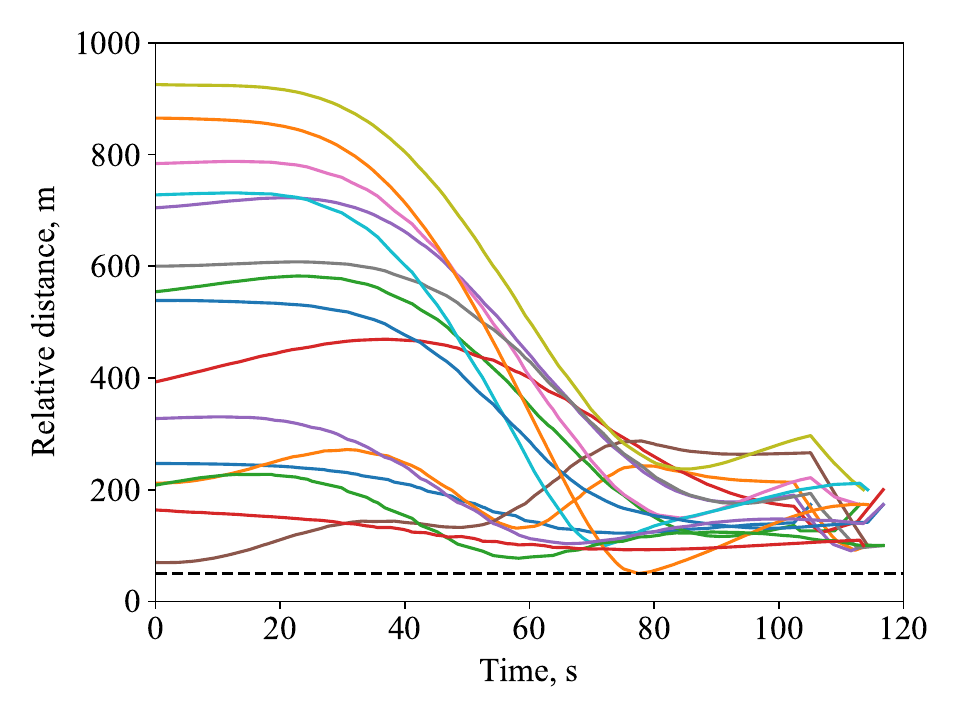}\\
	\caption{}
	\label{Fig8b}
\end{subfigure}
\caption{Collision avoidance trajectory for multiple parafoil; (a) 3-D trajectory (b) Relative distance between parafoils.}
\label{Fig8}
\end{figure*}

After the landing point allocation is determined, the flight trajectory of each parafoil system is decided. Through the trajectory check, two parafoils have a collision trajectory with each other, where the minimum distance is 32.89m. The trajectory replanning problem (\ref{equation: 25}) is solved by each parafoil in sequence, following the order of increasing $R$ value. The planned trajectories are shown in Fig. \ref{Fig8}, where the relative distances are kept above the minimum threshold $d_s$.

\begin{figure*}[!ht]
\begin{subfigure}[t]{.5\textwidth}
	\centering
	\includegraphics[width=14pc]{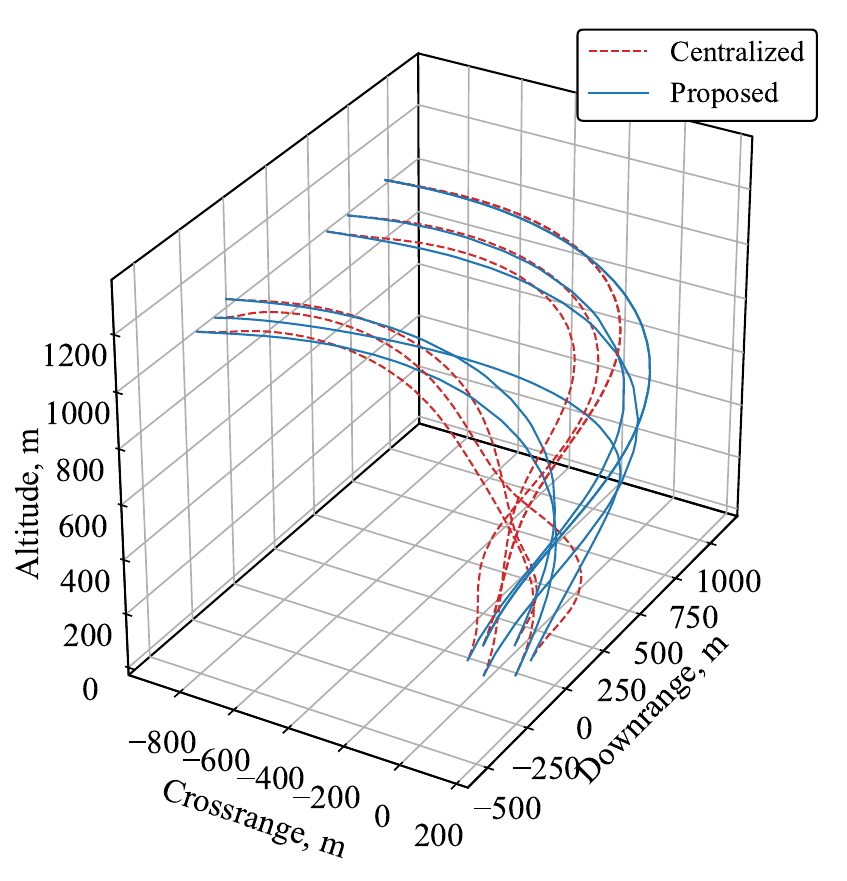}\\
	\caption{}
	\label{Fig9a}
\end{subfigure}
\begin{subfigure}[t]{.5\textwidth}
	\centering
	\includegraphics[width=16pc]{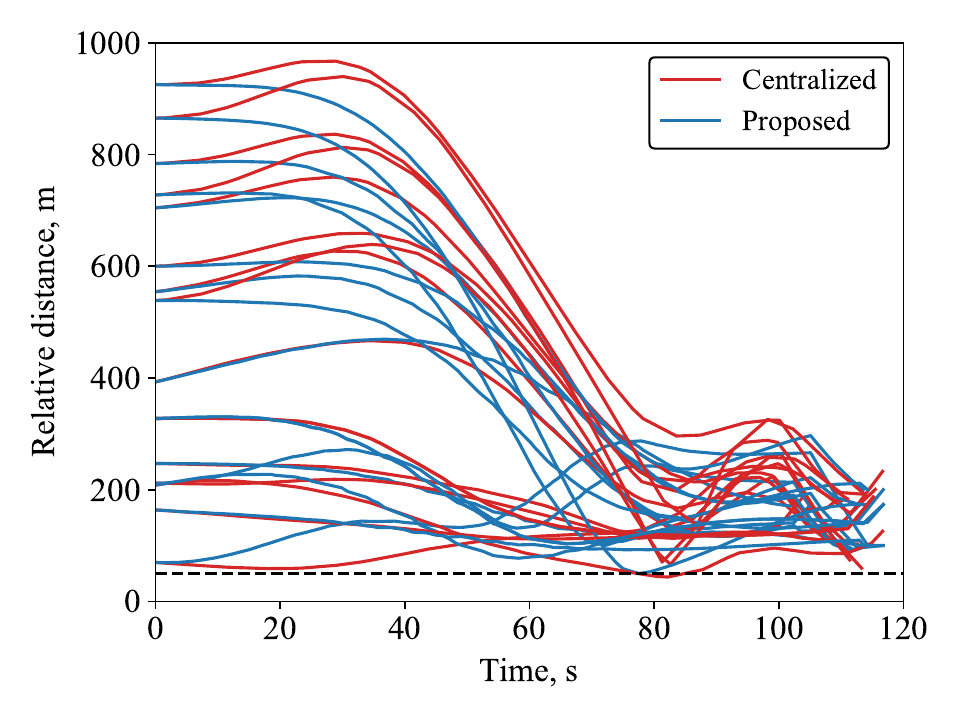}\\
	\caption{}
	\label{Fig9b}
\end{subfigure}
\caption{Comparison of trajectory planning framework; (a) 3-D trajectory (b) Relative distance between parafoils.}
\label{Fig9}
\end{figure*}

To reveal the advantage of the proposed collision-free trajectory planning algorithm (denoted as proposed), it is compared with simultaneously solving the multiple parafoil landing problem (denoted as centralized). Fig. \ref{Fig9a} visualizes the comparison of trajectories, and Fig. \ref{Fig9b} compares relative distance. The comparison results are summarized in Table 3. The proposed method finds different flight trajectories for the parafoils than that of the centralized framework, as the overall problem is considered in a distributed manner and collision avoidance trajectory is obtained from replanning. It should note that the proposed method consumes significantly less computation time owing to the decoupled problem formulation. Besides, the proposed framework obtains trajectories that strictly satisfy the minimum distance requirement, while the centralized framework fails to generate the collision-free trajectories. The reason is explained as follows. Take the example of \textit{i}th parafoil and \textit{j}th parafoil, the collision avoidance constraint in the centralized framework is formulated as $\left\| r_{R}^{i}\left( t_i \right) -r_{R}^{j}\left( t_j \right) \right\| ^2\geqslant \left( d_s \right) ^2$, where $t_i$ and $t_j$ are the discrete points for $t_i\in \left[ 0,t_{f}^{i} \right]$ and $t_j\in \left[ 0,t_{f}^{j} \right]$. $t_{f}^{i}$ and $t_{f}^{j}$ are the terminal time of each parafoil. Since the parafoil system is unpowered and has limited control over its vertical movement \cite{[22]yakimenko2015precision}, it is usually the case that $t_{f}^{i}\ne t_{f}^{j}$, which leads to $t_i \ne t_j$.

\begin{table}[!ht]
\caption{Comparison results of trajectory optimization frameworks.}
\label{Tab3}
\centering
\begin{tabular}{cccc}
\hline
Framework & Opt time, s & Min distance, m & Control energy\\
\hline
Centralized & 17.88 & 43.55 & 124.32\\
Proposed & 0.47 & 50.00 & 73.42\\
\hline
\end{tabular}
\end{table}

\subsection{Performance of moving horizon correction}
\begin{figure}[!ht]
\centering
\includegraphics[width=15pc]{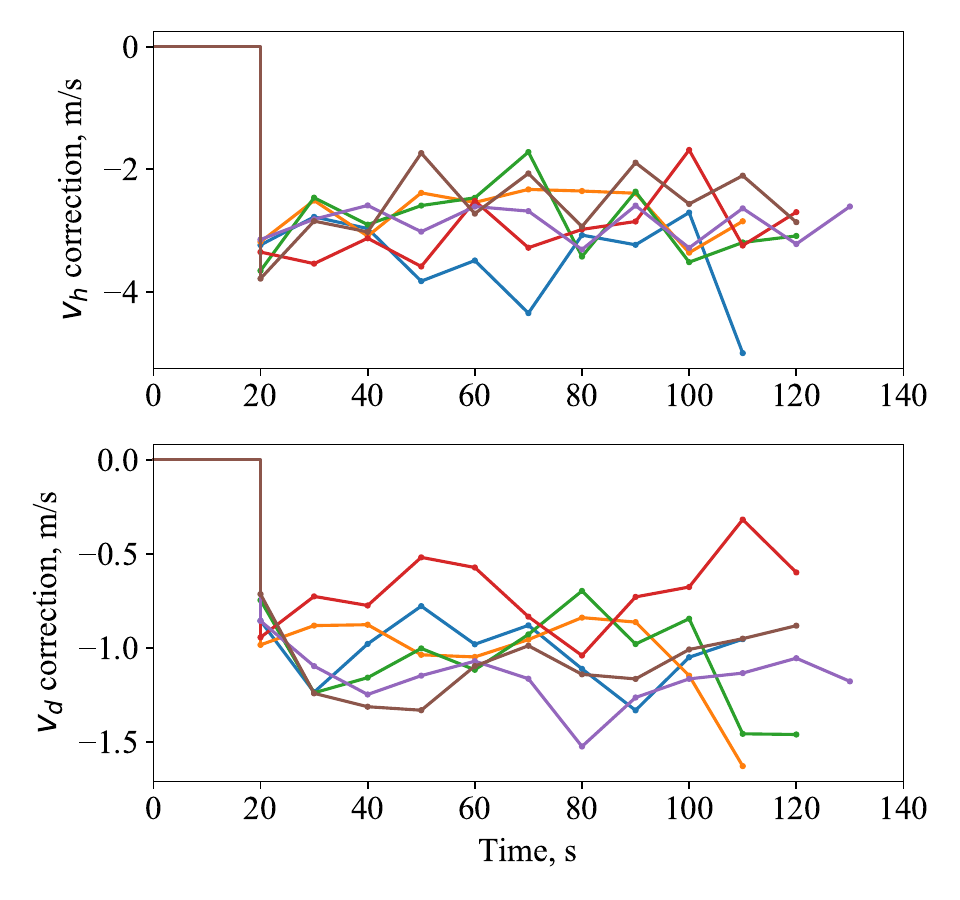}\\
\caption{Kinematic model correction result.}
\label{Fig10}
\end{figure}

\begin{figure}[!ht]
\centering
\includegraphics[width=15pc]{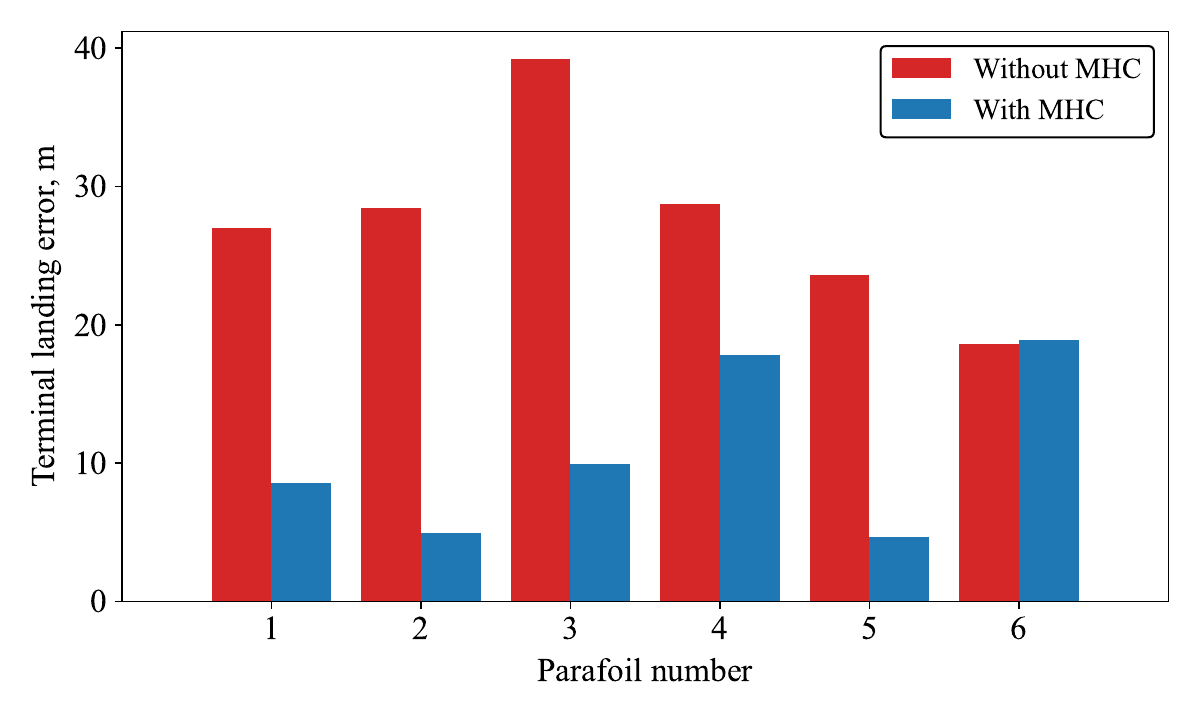}\\
\caption{Comparison of landing dispersion result.}
\label{Fig11}
\end{figure}

Fig. \ref{Fig10} shows the result of the moving horizon correction (MHC) algorithm, where the corrected horizontal and vertical velocity terms are added to the kinematic model to account for model inaccuracy. Fig. \ref{Fig11} compares the landing dispersion results with and without the kinematic model correction. It is shown that with the moving horizon correction algorithm, the terminal landing error is significantly reduced.

\subsection{Performance of coordinated guidance and control method}
\begin{figure*}[!ht]
\begin{subfigure}[t]{.5\textwidth}
	\centering
	\includegraphics[width=14pc]{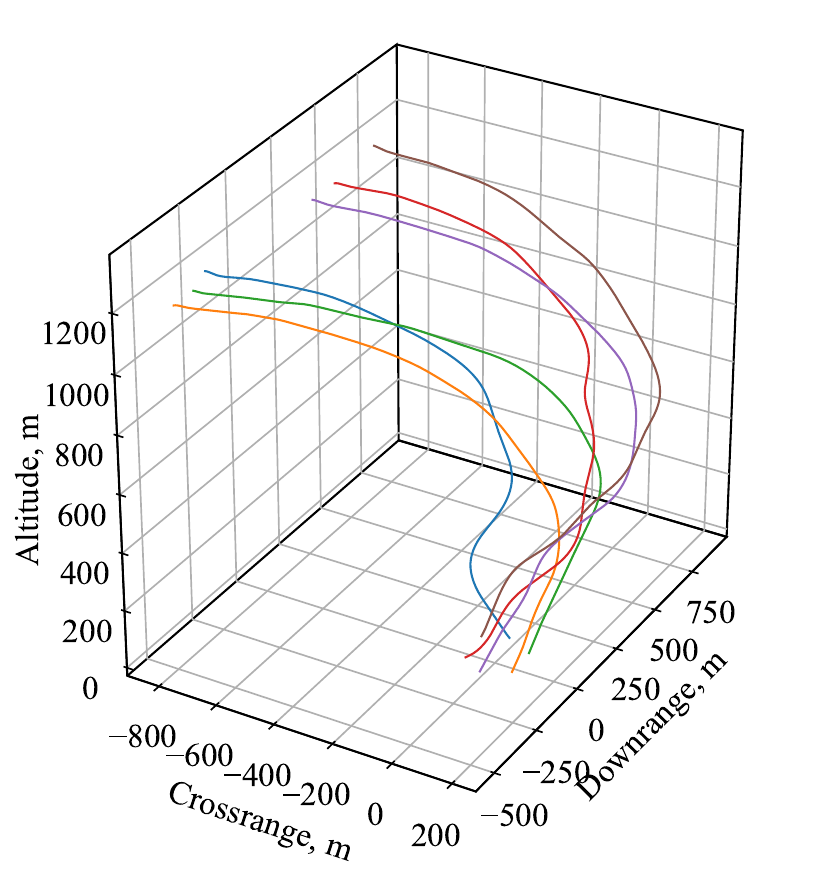}\\
	\caption{}
	\label{Fig12a}
\end{subfigure}
\begin{subfigure}[t]{.5\textwidth}
	\centering
	\includegraphics[width=16pc]{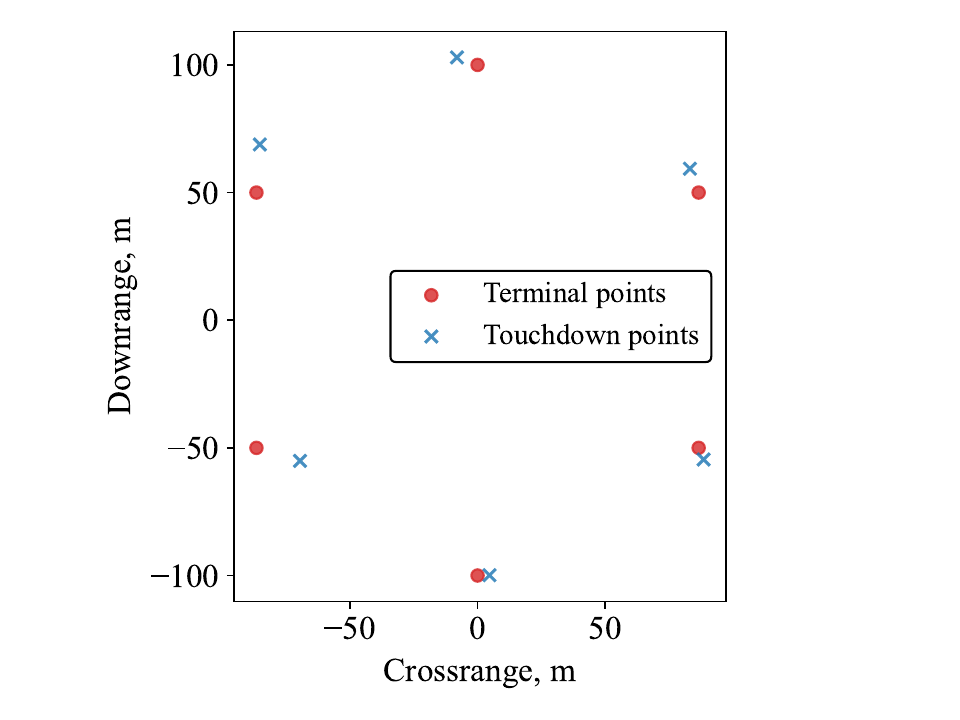}\\
	\caption{}
	\label{Fig12b}
\end{subfigure}
\caption{Closed-loop landing trajectory result; (a) 3-D trajectory; (b) Landing dispersion.}
\label{Fig12}
\end{figure*}

The closed-loop simulation results of the multiple parafoil are shown in Fig. \ref{Fig12}. The parafoils are coordinated to avoid collision and to land at the designated landing point. The maximum landing dispersion to the landing points is 18.89m, and the maximum distance to the landing area center is 109.59m.

\begin{figure}[!ht]
\centering
\includegraphics[width=15pc]{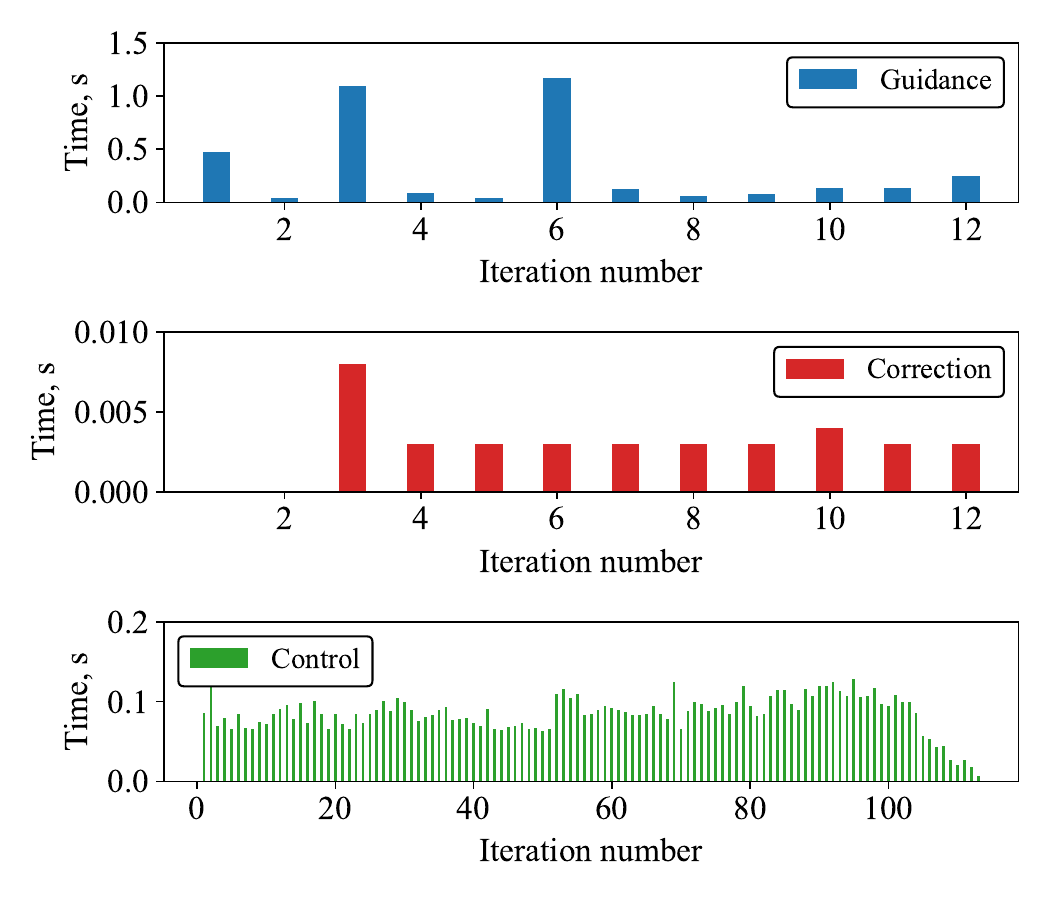}\\
\caption{Computation time of the proposed approach.}
\label{Fig13}
\end{figure}

The optimization problem computation time in each guidance and control loop is shown in Fig. \ref{Fig13}. The maximum computation time of the guidance trajectory is 1.17 seconds, and the maximum model correction time is 0.008 seconds. For the control horizon, the maximum NMPC solution time is 0.128 seconds.

\begin{figure}[!ht]
\centering
\includegraphics[width=15pc]{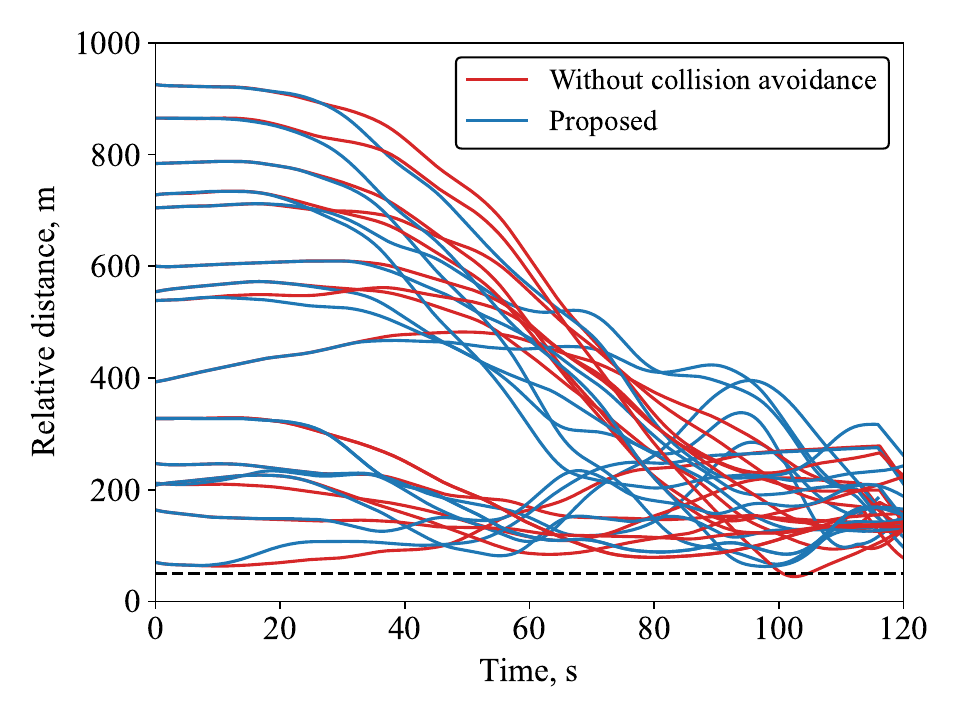}\\
\caption{Relative distance between parafoils in closed-loop simulation.}
\label{Fig14}
\end{figure}

The comparison of the relative distance with and without collision-free trajectory replanning is shown in Fig. \ref{Fig14}. With the proposed collision-free trajectory replanning algorithm, the minimum relative distance between parafoils increases from 44.37m to 62.53m, which ensures the safety of each parafoil.

\section{Conclusions}
\label{Sec5}
A coordinated guidance and control method has been proposed for multiple parafoil landing. With the NLP sensitivity analysis and the Hungarian algorithm, the landing points are allocated to each parafoil system by solving an assignment problem. Parafoils having the risk of collision are coordinated to replan the collision-free trajectory by solving decoupled trajectory optimization sub-problems. The kinematic model utilized in the trajectory planning process is updated by the moving horizon correction algorithm, and the generated reference trajectory is tracked by the NMPC algorithm.

Simulation results indicate that:
\begin{enumerate}
\item The landing points are efficiently evaluated via NLP sensitivity analysis, which consumes less computation time than repeatedly solving NLP problems.
\item The landing points are optimally assigned by the Hungarian algorithm, and the computation time is negligible.
\item The collision-free trajectory replanning algorithm ensures the minimum distance above the threshold, and requires less computational effort than the centralized framework.
\item The moving horizon correction algorithm reduces the landing error of each parafoil by updating the kinematic model parameters.
\item The proposed coordinated guidance and control method is capable of realizing multiple parafoil landing, and its computation time satisfies the online demand.
\end{enumerate}

In our future work, the reformulation and reduction of collision avoidance constraints will be considered to reduce the path constraint complexity. Besides, collision probability estimation, robust collision-free trajectory planning, and constant wind field estimation will also be considered to extend the proposed framework to scenarios with changing wind fields and strong disturbances. On this basis, the proposed guidance and control framework will be verified through experiments.

\bibliographystyle{IEEEtran}
\bibliography{main}

\end{document}